\PassOptionsToPackage{table}{xcolor}
\documentclass[11pt]{article}

\usepackage[final]{acl}

\usepackage{times}
\usepackage{latexsym}

\usepackage[T1]{fontenc}

\usepackage[utf8]{inputenc}

\usepackage{microtype}

\usepackage{inconsolata}

\usepackage{graphicx}

\usepackage{subcaption}
\usepackage{booktabs} 

\usepackage{amsmath}
\usepackage{amssymb}
\usepackage{mathtools}
\usepackage{amsthm}

\usepackage{pifont}
\usepackage{multirow}
\usepackage{graphicx}
\usepackage{pdflscape}
\usepackage{enumitem}
\usepackage[most]{tcolorbox}
\tcbuselibrary{listings, breakable}
\usepackage{float}
\usepackage{comment}
\usepackage{capt-of}
\usepackage{xcolor}
\usepackage{algorithm}
\usepackage{algorithmic}
\usepackage{textcomp}
\usepackage{upquote}
\usepackage{makecell}
\usepackage{array}
\usepackage{placeins}
\usepackage{newfloat}

\definecolor{claritybg}{HTML}{FFF2CC} 
\newcolumntype{Y}{>{\columncolor{claritybg}}c}

\tcbset{
    mymultipagebox/.style={
        enhanced,
        breakable,         
        colback=gray!10,   
        colframe=black,    
        left=1mm, right=1mm, top=1mm, bottom=1mm,
        boxrule=0.5pt,
        arc=2mm,
    }
    }

\newcommand{\wheelsym}{\raisebox{-0.2em}{\includegraphics[height=1.2em]{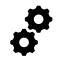}}}
\newcommand{\llmsym}{\raisebox{-0.2em}
{\includegraphics[height=1.2em]{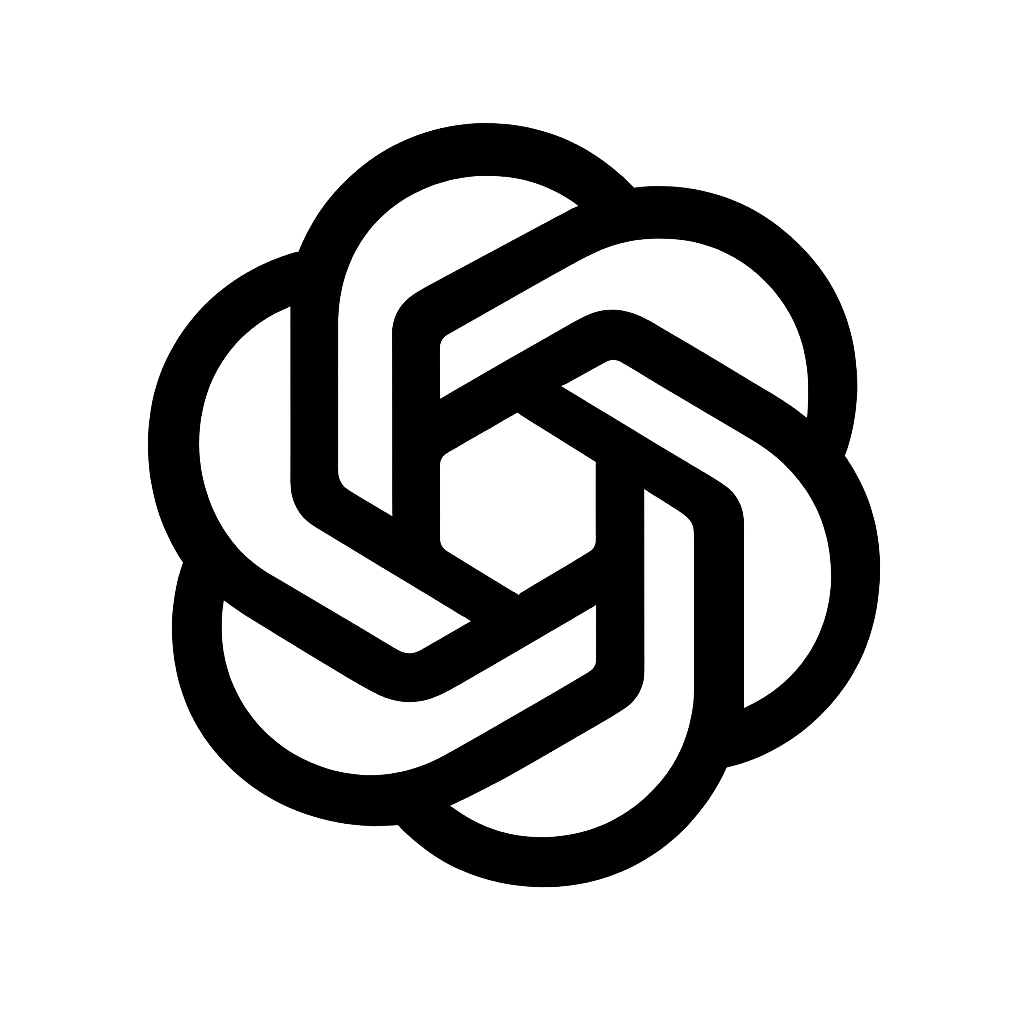}}}

\usepackage{caption}
\usepackage[capitalize,noabbrev]{cleveref}

\theoremstyle{plain}

\theoremstyle{definition}

\theoremstyle{remark}

\usepackage[textsize=tiny]{todonotes}

\title{CLARITY: A Framework and Benchmark for 
Conversational Language Ambiguity and Unanswerability in Interactive NL2SQL Systems}

\author{
  Tabinda Sarwar,
  Farhad Moghimifar,
  Cong Duy Vu Hoang\thanks{Corresponding author: vu.hoang@oracle.com},
  Xiaoxiao Ma,
  \\
  \bf Shawn Chang Xu,
  Fahimeh Saleh,
  Poorya Zaremoodi,
  \\
  \bf Avirup Sil,
  Katrin Kirchhoff
\\
  \normalfont Oracle Corporation
}

\begin{document}
\maketitle

\begin{abstract}

NL2SQL systems deployed in industry settings often encounter ambiguous or unanswerable queries, particularly in interactive scenarios with incomplete user clarification. Existing benchmarks typically assume a single source of ambiguity and rely on user interaction for resolution, overlooking realistic failure modes.

We introduce \textsc{Clarity}, a framework for automatically generating an NL2SQL benchmark with multi-faceted ambiguities and diverse user behaviors across both single- and multi-turn settings. Using a constraint-driven pipeline, \textsc{Clarity} transforms executable SQL into ambiguous queries, augmented with grounded conversational continuations and schema-level metadata.

Empirical evaluation on Spider and BIRD shows that leading NL2SQL systems, including those based on strong LLMs, suffer significant performance degradation under multi-faceted ambiguity. While these systems often detect ambiguity, they struggle to accurately localize and resolve the underlying schema-level sources. Our results highlight the need for more robust ambiguity detection and resolution in industry-grade NL2SQL systems.

\end{abstract}

\section{Introduction} \label{sec:intro}

Natural language to SQL (NL2SQL) systems are increasingly being deployed as production interfaces for non-expert users to query structured data \cite{shi2025survey, hong2025next, liu2025survey}. Despite substantial progress driven by large language models (LLMs), real-world usage continues to present a fundamental challenge: user queries are frequently ambiguous or unanswerable, and these failure modes are amplified in interactive settings where user clarifications may be incomplete, vague, or even irrelevant \cite{huang2025exploring, NEURIPS2023_72223cc6, talaei2024chess, lee2025mcs, chen2025reliable}. 
In practice, a single query often contains multiple interacting sources of ambiguity, covering both schema columns and values, yet most existing benchmarks and evaluation protocols assume a single ambiguity per query and cooperative users who provide clean clarifications \cite{ambisql, zhao2024sphinteract, vaidya2025odin, saparina2025disambiguate}.

Previous work has addressed ambiguity and unanswerability in NL2SQL, mainly in single-turn or simplified interactive settings as shown in Table~\ref{tab:sota}. Benchmarks such as AMBROSIA \cite{ambrosia}, AmbiQT \cite{bhaskar2023benchmarking}, NoisySP \cite{wang2023know} and SQUAB \cite{squab} target isolated ambiguity phenomena, while PRACTIQ \cite{practiq}, MMSQL \cite{mmsql}, and BIRD-INTERACT \cite{birdinteract} extend the evaluation to multi-turn scenarios. However, these benchmarks rely largely on instance-level labels (e.g., ambiguous vs. unanswerable) and do not assess whether systems correctly localize the schema elements responsible for uncertainty. Consequently, high ambiguity-detection accuracy may mask failures in schema-level identification and resolution. We argue that fine-grained, schema-grounded evaluation enables more diagnostic assessment of NL2SQL robustness, particularly under multi-faceted ambiguity and realistic interaction settings.

\begin{figure}[!t]
    \centering
    \includegraphics[width=\columnwidth]{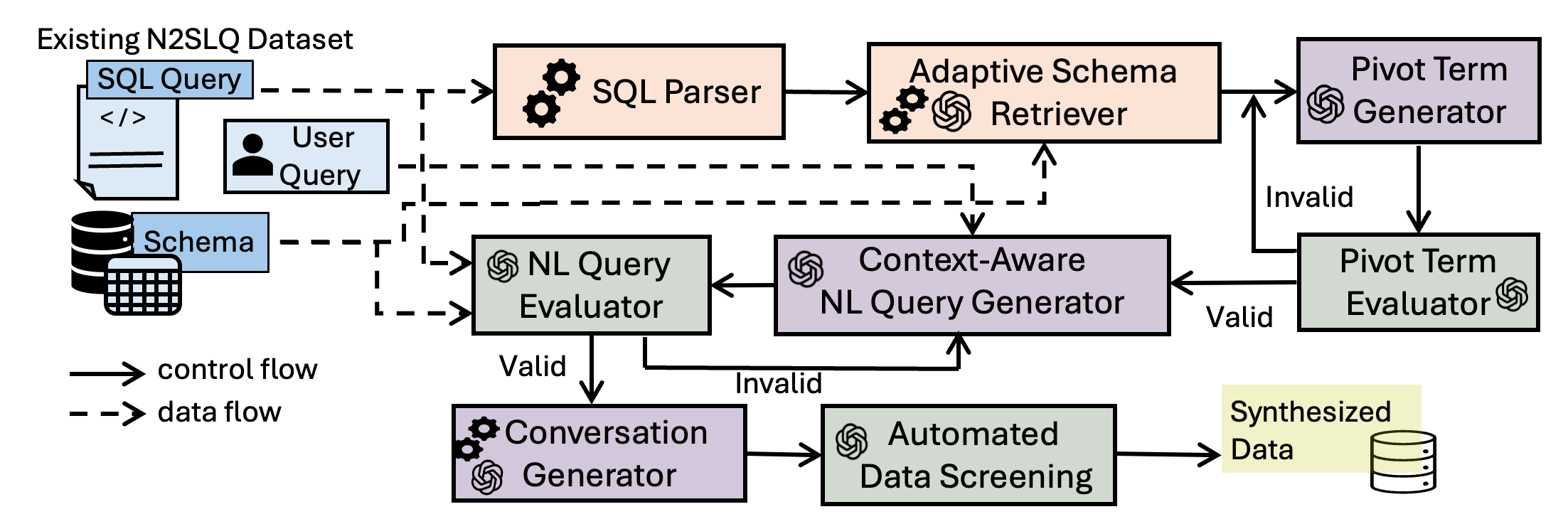}
    \caption{\textbf{Overview of the \textsc{CLARITY} framework}. The symbols \wheelsym\ and \llmsym\ denote rule-based and LLM-based components, respectively.}
    \label{fig:framework}
\end{figure}

\begin{table*}[!t]
\centering
\small
\resizebox{\textwidth}{!}{
\begin{tabular}{l l c c c c c c c Y}
\hline
\textbf{Schema Entity} & \textbf{Use Case} & \textbf{AMBROSIA} & \textbf{AmbiQT} & \textbf{NoisySP} & \textbf{SQUAB} & \textbf{PRACTIQ} & \textbf{MMSQL} & \textbf{BIRD-INTERACT} & \textbf{CLARITY (OURS)} \\
\hline
\textbf{Type of Data} & {-} & {Single turn}  & {Single turn}  & {Single turn}  & {Single turn}  & {Multi turn}  & {Multi turn} &  {Multi turn}   & {Single and multi turn} \\
\hline
\multirow{4}{*}{Column} & {Unifacet Amb} & {\ding{51}} & {\ding{51}} & {\ding{51}} & {\ding{51}} & {\ding{51}} & {\ding{51}} & {\ding{51}}  & {\ding{51}} \\
                        & {Unifacet Unans} & {\ding{51}} & {\ding{51}} & {\ding{51}} & {\ding{51}} & {\ding{51}} & {\ding{51}} & {\ding{51}}  & {\ding{51}}\\
                        & {Multifacet Amb} & {\ding{55}} & {\ding{55}} & {\ding{55}} & {\ding{55}} & {\ding{55}} & {\ding{55}} & {\ding{55}}  & {\ding{51}}\\
                        & {Multifacet Unans} & {\ding{55}} & {\ding{55}} & {\ding{55}} & {\ding{55}} & {\ding{55}} & {\ding{55}} & {\ding{55}}  & {\ding{51}}\\
\multirow{4}{*}{Value}  & {Unifacet Amb} & {\ding{55}} & {\ding{55}} & {\ding{51}} & {\ding{55}} & {\ding{51}} & {\ding{51}} & {\ding{55}}  & {\ding{51}}\\
                        & {Unifacet Unans} & {\ding{55}} & {\ding{55}} & {\ding{51}} & {\ding{55}} & {\ding{51}} & {\ding{55}} & {\ding{55}}  & {\ding{51}}\\
                        & {Multifacet Amb} & {\ding{55}} & {\ding{55}} & {\ding{55}} & {\ding{55}} & {\ding{55}} & {\ding{55}} & {\ding{55}}  & {\ding{51}}\\
                        & {Multifacet Unans} & {\ding{55}} & {\ding{55}} & {\ding{55}} & {\ding{55}} & {\ding{55}} & {\ding{55}} & {\ding{55}}  & {\ding{51}}\\
\hline
\end{tabular}
}
\caption{Comparison with related works on ambiguity and unanswerable questions in interactive NL2SQL systems.}
\label{tab:sota}
\end{table*}


To address this limitation, we introduce \textbf{C}onversational \textbf{L}anguage \textbf{A}mbiguity and unanswe\textbf{R}ability  in \textbf{I}nterac\textbf{T}ive NL2SQL s\textbf{Y}stems (\textsc{Clarity}), a novel unified benchmark for evaluating ambiguous and unanswerable NL2SQL queries under realistic interactions. \textsc{Clarity} automatically synthesizes NL2SQL instances with multiple interacting ambiguity and unanswerability sources, supports both single- and multi-turn interactions, and models diverse user clarification behaviors, including partially helpful and unhelpful responses that commonly arise in production systems. Crucially, \textsc{Clarity} augments each instance with fine-grained, schema-grounded metadata, including span-level pivot terms and candidate schema target groups, enabling more precise evaluation of NL2SQL systems and pinpointing areas most in need of improvement, a capability not supported by existing benchmarks. 
This is essential for building robust production NL2SQL systems, where users depend on real-time analysis, identifying the precise causes of failures is therefore essential.

\textsc{Clarity} generates data without human annotation, enabling low-effort in-house creation of ambiguous and unanswerable variants from enterprise datasets. Instantiating \textsc{Clarity} on Spider \cite{yu-etal-2018-spider} and BIRD \cite{birddataset} shows that state-of-the-art LLMs degrade under multi-facet ambiguity and that high ambiguity-detection accuracy does not guarantee correct schema localization or resolution.

We make the following contributions:
\begin{itemize}[nosep]
  \item \textbf{\textsc{Clarity} Benchmark:} We introduce a diagnostic benchmark that systematically captures \emph{multi-facet} ambiguity and unanswerability in both single-turn and multi-turn NL2SQL (\cref{definition}). Although our approach leverages LLMs, its main contribution is a scalable, schema-aware framework for controlled generation of ambiguous and unanswerable NL2SQL data. 

  \item \textbf{Schema-Grounded, Fine-Grained Annotations.} Instances are augmented with span-level pivot terms and corresponding schema target groups, enabling schema-aware evaluation beyond coarse instance-level labels.

  \item \textbf{Empirical Diagnosis of LLM Limitations.} Extensive experiments show that state-of-the-art LLMs frequently detect ambiguity without correctly localizing the underlying schema elements, with performance degrading sharply under multi-facet ambiguity, failure modes not exposed by existing benchmarks.
\end{itemize}

\section{Benchmark Construction Methodology}
\label{sec:method}

\subsection{Problem Statement}

Let $\mathcal{D} = \{(u_i, q_i, \mathcal{S}_i)\}_{i=1}^{N}$ denote an NL2SQL dataset, where $u_i$ is a natural language query, $q_i$ is its corresponding SQL query, and $\mathcal{S}_i$ is the associated database schema.

Our goal is to construct a synthetic dataset $\mathcal{D}^{A/U}$ consisting of single-turn and multi-turn NL2SQL instances in which the initial user query may contain one or more sources of \emph{ambiguity} or \emph{unanswerability} (A/U) with respect to the schema. Each instance is associated with:
\\
(i) a set of A/U modes
\[
M \subseteq \{\texttt{col\_amb}, \texttt{val\_amb}, \texttt{col\_unans}, \texttt{val\_unans}\},
\]
(ii) a conversation type
\[
R \in \{\texttt{helpful}, \texttt{partial}, \texttt{unhelpful}\},
\]
(iii) a set of \emph{pivot terms} $\{p_j\}$, and
(iv) a corresponding set of schema-grounded column or value groups $\{G_j\}$.

The pivot terms and target groups provide fine-grained supervision for ambiguity detection and localization.

Given an instance $(u, q, \mathcal{S}, M, R, \{p_j\}, \{G_j\})$, the framework generates a multi-turn interaction
\[
C = \{(r_1, t_1), \dots, (r_T, t_T)\},
\]
where $r_t$ and $t_t$ denote user and agent utterances at turn $t$, respectively, and $r_1 = u$ is the initial (potentially ambiguous or unanswerable) query. For single-turn settings, $C$ reduces to $\{(r_1, t_1)\}$, and $r_1$ alone constitutes the generated A/U query.

\subsection{Framework Overview}

\textsc{Clarity} is a modular, end-to-end framework, illustrated in \cref{fig:framework} and Algorithm~\ref{alg:au-framework}. We outline the main procedure here, with detailed descriptions of each module provided in Appendix~\ref{appendix:pseudocode}. For \textsc{Clarity}, we omit instance indices and describe the procedure for a single input $(u, q)$.
\begin{algorithm}[!t]
\caption{The \textsc{Clarity} Framework}
\footnotesize
\label{alg:au-framework}
\begin{algorithmic}[1]
\REQUIRE schema $S$, NL query $u$, SQL query $q$, mode $m$, response type $R$, number of cases $K$
\ENSURE conversation $C$, validation flag $v \in \{\textsc{True}, \textsc{False}\}$

\STATE $t \gets \textsc{SQLParser}(q, m, K)$ \hfill // $t[1..K]$: targets

\FOR{$i = 1$ to $K$}
    \STATE $G[i] \gets \textsc{AdaptiveSchemaRetriever}(t[i], S, m)$
    \STATE $f_p \gets \emptyset$
    \REPEAT
        \STATE $p[i] \gets \textsc{PivotGenerator}(t[i], G[i], m, f_p)$
        \STATE $(f_p, ok) \gets \textsc{PivotEvaluator}(p[i], G[i], m)$
    \UNTIL{$ok$}
\ENDFOR

\STATE $f_q \gets \emptyset$
\REPEAT
    \STATE $r \gets \textsc{QueryGenerator}(u, q, p, t, f_q)$ \hfill // rewritten query
    \STATE $(f_q, ok) \gets \textsc{QueryEvaluator}(r, S, m, q, p, t)$
\UNTIL{$ok$}

\STATE $C \gets \textsc{ConversationGenerator}(r, R)$
\STATE $v \gets \textsc{AutomatedDataScreening}(C, p, t, S, m)$

\STATE \textbf{return} $C, v$
\end{algorithmic}
\end{algorithm}

\paragraph{SQL Parser}

Similar to \cite{practiq}, we developed an SQL Parser which analyzes the input query $q$ to extract (i) the set of referenced columns $\mathcal{C}$ and (ii) the set of column--value pairs $(\mathcal{C}, \mathcal{V})$ induced by the predicate structure. It then samples a set of $K$ A/U targets
\[
\mathcal{T} = \{t_j\}_{j=1}^{K},
\]
where each target $t_j$ corresponds either to a column $(m_j, c_j)$ or to a column--value pair $(m_j, c_j, v_j)$ with mode $m_j \in M$. The value of $K$ controls whether the uni- or multi-facet A/U cases. Sampling is performed without replacement to avoid degenerate overlaps.

\begin{table}[t]
\label{definition}
\centering
\small
\begin{tabular}{p{0.27\columnwidth} p{0.68\columnwidth}}
\hline
\textbf{Concept} & \textbf{Definition} \\
\hline

\textbf{Ambiguity} &
A term can be linked to two or more schema entities (column names or values). Subcategories include:
\textit{Lexical} (shared tokens or surface forms) and
\textit{Semantic} (shared or closely related meanings). \\
\hline

\textbf{Unanswerable} &
A term cannot be linked to any schema entity (column names or their values). \\
\hline

\textbf{Column Use Case} &
The A/U case is associated with a schema column. \\
\hline

\textbf{Value Use Case} &
The A/U case is associated with a column value. \\
\hline

\textbf{Facet-based Use Case} &
\textit{Uni-facet:} The query contains a single A/U case.
\textit{Multi-facet:} The query contains two or more A/U cases. \\
\hline

\textbf{Single-turn Instance} &
An ambiguous NL query that maps to multiple valid SQL queries, or to none if unanswerable. \\
\hline

\textbf{Multi-turn Instance} &
A user--agent dialogue in which turns aim to resolve ambiguity or unanswerability in the NL query. \\
\hline

\end{tabular}
\caption{Definitions of ambiguity and unanswerable (A/U) use cases. Examples are provided in \cref{definition_examples}.}
\end{table}

\paragraph{Adaptive Schema Retriever}

To handle complex datasets \cite{kannan2025high}, we introduce an adaptive schema retriever. For each target $t_j \in \mathcal{T}$, it constructs a target space $\mathcal{X}_j$ of relevant schema entities—columns for column-based modes and column values for value-based modes.

For ambiguity modes, it retrieves a candidate group $G_j \subset \mathcal{X}_j$ of lexically or semantically similar elements using LLM-based, dense, or hybrid retrieval, with the strategy and group size adapted to the schema scale and mode. Lexical similarity is computed via token overlap, while semantic similarity is measured using cosine similarity between sentence embeddings; hybrid retrieval combines both signals.

For unanswerable modes, no candidate group is formed. Instead, the full target space serves as a negative reference, ensuring that the generated pivot term does not match any valid schema element.

To distinguish lexical from semantic ambiguity, we apply a two-stage filtering procedure: lexical candidates are first identified via token-level overlap, and semantic candidates are then retrieved using embedding similarity under a lexical-overlap constraint. This separation yields ambiguity cases driven primarily by surface-form versus conceptual similarity.


\paragraph{Pivot Term Generator}

The Pivot Term Generator synthesizes a set of pivot terms $\{p_j\}_{j=1}^{K}$, one per A/U target using LLM \cite{nadua2025synthetic}. For each target $t_j$, it conditions on the retrieval context $(\mathcal{X}_j, G_j)$ and a global exclusion set of previously generated pivot terms. If $m_j$ is an ambiguity mode, the LLM is instructed to generate a term $p_j$ that is compatible with multiple elements in $G_j$. If $m_j$ is an unanswerable mode, it generates a term that is dissimilar to all elements in $\mathcal{X}_j$. Prompts explicitly prohibit generating terms that already exist in the schema, or previously generated pivot terms, preventing leakage and collisions. $\{p_j\}$ and $\{G_j\}$ serve as \textbf{\textit{metadata}} for A/U detection and resolution.

\paragraph{Pivot Term Evaluator}

Each pivot term is independently validated by multiple LLM judges \cite{verga2024replacing} against a set of boolean constraints: (i) conformity to the intended A/U mode, (ii) compatibility with the corresponding target group $G_j$ (for ambiguity modes), and (iii) absence from the schema $\mathcal{S}$. Any violation must be accompanied by a short justification, which is fed back to the Pivot Term Generator for correction. A pivot term proceeds only if all judges unanimously pass all checks.

\paragraph{Context-Aware NL Query Generator}

This module integrates pivot terms into $u$ to produce an A/U query $r_1$, enforcing: (i) correct semantic placement of each pivot term based on $q$; (ii) semantic consistency with $q$ under the intended resolution; and (iii) preservation of the syntactic structure, intent, and writing style of $u$. This controlled transformation avoids fully synthetic generation, reducing unrealistic phrasing and mitigating potential LLM biases. A context-suppression constraint prevents unintended disambiguation, and failed generations are iteratively refined using evaluator feedback.


\paragraph{NL Query Evaluator}

The generated query $r_1$ is evaluated by multiple LLM judges to ensure that: (i) all pivot terms appear exactly once, (ii) the query satisfies the formal A/U definition under $\mathcal{S}$, and (iii) resolving the pivot terms according to $\mathcal{T}$ yields a query consistent with $q$. Only queries that unanimously pass all criteria are retained.

\paragraph{Conversation Generator}

This module extends $r_1$ into a multi-turn interaction. It first produces an agent clarification utterance targeting $p_j$, which is then diversified by an LLM. Conditioned on the sampled $R$, the LLM generates alternating user and agent turns, producing a conversation
\[
C = \{(r_t, t_t)\}_{t=1}^{T}.
\]
 For ambiguity cases, user responses are integrated into $q$ to produce a final SQL query; for unanswerable cases, the agent explicitly states that the query cannot be executed.

\paragraph{Automated Data Screening}

Inspired by \cite{talaei2024chess}, we developed an Automated Data Screening module to evaluate each conversation against test cases designed to comply with the A/U specification under $\mathcal{S}$. Using a mixture of LLM judges, each conversation is validated against these test cases, and instances failing majority voting are discarded. 

\section{Data Statistics}


We construct A/U datasets on the development and test splits of Spider \cite{yu-etal-2018-spider} and BIRD \cite{birddataset}, two large-scale, human-curated NL2SQL benchmarks that pair natural language queries with executable SQL across diverse schemas and domains. Using the \textsc{Clarity} framework, we generate data without additional human annotation or schema augmentation. GPT-5 is used for generation, while GPT-4o, LLaMA-3.3, and Grok-3 Mini Fast serve as evaluators. We target multi-facet settings by introducing up to two A/U instances per query.

In total, we generate 6,392 (Spider) and 3,487 (BIRD) single-turn instances, and 12,098 (Spider) and 7,171 (BIRD) multi-turn instances across four conversation types. Dataset statistics are summarized in \cref{statistics}. The resulting distribution reflects both the structural properties of the source datasets and the pipeline’s strict validation criteria. Multi-facet samples are fewer due to the limited prevalence of multi-column SQL queries and stringent pivot- and query-level validation (see \cref{tab:failure_cases} for examples). Despite their lower frequency, multi-facet instances exhibit higher semantic validity and diagnostic value.

To prevent unbounded regeneration, the evaluator phase is capped at five iterations. Example outputs are provided in \cref{tab:examples}.

\begin{table}[t]
  \centering
  \label{statistics}
  \resizebox{\columnwidth}{!}{
    \begin{tabular}{l c c c c c}
     \toprule
      \textbf{Dataset} & \textbf{Use Case} & \makecell{\textbf{Uni-}\\\textbf{Facet}} & \makecell{\textbf{Multi-}\\\textbf{Facet}} & \makecell{\textbf{Multi-}\\\textbf{Turn}} & \makecell{\textbf{Acc}\\\textbf{Human Ann}}\\
        \midrule
      \multirow{6}{*}{Spider} & Lexical Amb & 490 & 39 & 2,645 & 100\\
                               & Semantic Amb & 612 & 57 &  3,345 & 93.7 \\
                               & Value Amb & 297 & 9 & 1,233 & 93.7\\
                               & Unans Column & 1,859 & 1,360 & 3,219 & 100 \\
                               & Unans Value & 1,456 & 200 &  1,656 & 100\\
          \midrule
        \multirow{6}{*}{Bird} & Lexical Amb & 145 & 13 & 790 & 100\\
                               & Semantic Amb & 260 & 32 & 1,460 & 100\\
                               & Value Amb & 421 & 50 & 2,355 & 81.2\\
                               & Unans Column & 1,071 & 925 & 1,996 & 100\\
                               & Unans Value & 477 & 93 & 570 & 100 \\
      \bottomrule
    \end{tabular}
      }
    \caption{Data Statistics. ``\textit{Acc Human Ann}'' represents the accuracy of human validation on the sampled dataset.}
  \vskip -0.1in
\end{table}

To assess whether the generated dataset preserves key linguistic characteristics, we compare it against the original Spider and BIRD datasets. For BIRD, generated queries are longer on average (17.47 vs.\ 14.93 tokens) and exhibit broader lexical coverage (3,720 vs.\ 2,301 unique types). Similarly, for Spider, generated queries are slightly longer (13.69 vs.\ 12.37 tokens) and show increased lexical diversity (4,664 vs.\ 893 types).

Across both datasets, the distribution of query intents is largely preserved, with retrieval and counting intents accounting for approximately 92\% of queries. At the same time, the generated data introduces stronger signals of constraint and aggregation; for instance, the frequency of comparative markers in BIRD increases from 7.2\% to 19.9\%, reflecting the intended increase in ambiguity and reasoning complexity.

Overall, these results indicate that \textsc{Clarity} preserves realistic query characteristics while systematically enriching linguistic cues associated with ambiguity and unanswerability.



\section{Evaluation Tasks and Metrics}
\label{sec:eval_tasks}

\paragraph{Task.}
We define evaluation tasks for both single- and multi-turn instances. For value-level A/U, NL2SQL evaluation would require exposing full database contents to the model, which is impractical; context-offloading approaches that could mitigate this are beyond the scope of this work. We therefore exclude value-based A/U categories from this evaluation. Evaluation prompts are provided in Appendix~\ref{appendix:eval_appendix}.

\paragraph{Metrics.}
For Spider, we report Execution Accuracy (EA), Strict Exact Match (SEM), and Lenient Exact Match (LEM) \cite{li2024dawn} to evaluate structural correctness. For BIRD, which contains more complex SQL constructs, we report only EA. For unanswerable cases, the only valid output is \texttt{Null}.

\subsection{Human Validation}
\label{sec:human_ann}

Although the \textsc{Clarity} benchmark is generated fully autonomously, we conduct a human validation study to verify adherence to the formal A/U definitions in \cref{definition}. Three expert annotators with substantial experience in databases and NL2SQL independently evaluated a stratified sample, making binary judgments on whether each instance satisfies its assigned A/U definition, guided by examples distinguishing different A/U types. We sample 8 instances per A/U category from both Spider and BIRD, yielding a balanced set of 448 instances covering all use cases. Validation accuracy is computed via majority vote across annotators.

\subsection{SQL Prediction Under Ambiguity}

Ambiguous NL queries may admit multiple valid SQL interpretations, whereas unanswerable queries admit none. Accordingly, we design distinct evaluation protocols for single-turn and multi-turn settings.

\paragraph{Single-Turn Prediction.}
Models are prompted to enumerate all valid SQL interpretations of a given ambiguous query. Performance is evaluated using:
(i) Strict Exact Match (SEM), which requires the predicted set of SQL queries to exactly match the ground-truth set; and
(ii) Lenient Exact Match (LEM), which requires at least one correct SQL query to be generated.

SEM is particularly informative in multi-faceted settings, as it measures whether the model recovers the full set of valid interpretations.

\paragraph{Multi-Turn Prediction.}
Models are provided with the full conversation (excluding the final SQL) and must produce a single resolved SQL query. Since ambiguity is resolved through interaction, we evaluate using Execution Accuracy (EA), which measures whether executing the predicted SQL yields the same result as the ground-truth query.

This setup evaluates two complementary capabilities: (1) enumerating alternative interpretations in single-turn scenarios, and (2) resolving ambiguity through dialogue.

Following prior work \cite{li2024dawn}, we report SEM and LEM on Spider to assess structural correctness, and EA on the more complex BIRD dataset. For unanswerable queries, the only valid output is \texttt{null}. We report macro-averaged accuracy for cross-category comparison.

\subsection{Ambiguity Detection}

We evaluate ambiguity detection on single-turn instances using AmbiSQL \cite{ambisql}. Predicted ambiguities are compared against the ground-truth metadata from \textsc{Clarity}. We report \emph{Detection Accuracy (DA)}, which measures whether ambiguity is correctly identified, and \emph{Match Accuracy (MA)}, which assesses whether the predicted ambiguity aligns with the underlying schema elements. We further use our generated instances as few-shot exemplars to evaluate their impact on AmbiSQL’s performance.

\section{Results and Discussion}


\cref{statistics} summarizes the human annotation results. A 10–20\% accuracy drop is observed only for \texttt{multi val\_amb} cases; all other A/U categories show consistently high agreement. One additional error stems from a syntax issue in an underlying SQL query for \texttt{sem\_amb}, reflecting dataset-specific artifacts of the data-driven generation process. During generation, multiple evaluators are used to filter erroneous instances (see \cref{tab:failure_cases}), ensuring adherence to the A/U definitions and contributing to the overall high annotation accuracy.

We benchmark five strong LLMs: one open-source model (LLaMA-3.3) and four closed-source models (GPT-4o, GPT-4.1, GPT-5, and Grok-3 Mini Fast). For all models, we set \texttt{temperature = 0.0} and \texttt{max\_tokens = 2048}; for GPT-5, \texttt{reasoning\_effort} and \texttt{verbosity} are additionally set to \texttt{medium}. Model parameters are not tuned, and no prompt optimization is performed. Using zero-shot prompting as a baseline, we evaluate performance on the full \textsc{Clarity} benchmark. Prompts from \cite{ambrosia} and \cite{ambisql} are used without task-specific adaptation.

Under zero-shot prompting, LLMs struggle to enumerate multiple SQL interpretations for column-level ambiguity (\cref{tab:exact_match_ex}, \cref{tab:exact_match}). Low SEM indicates missed valid interpretations, while higher LEM reflects partial coverage (i.e., multiple but incomplete outputs).

We evaluate two few-shot strategies by randomly sampling two exemplars per \textsc{Clarity} ambiguity case (excluded from the test set). Each exemplar consists of an ambiguous NL query paired with multiple SQL queries. We consider: (i) \textit{without metadata}, and (ii) \textit{with metadata}, where metadata includes pivot-term and target-group annotations. Metadata is used only in-context to aid ambiguity detection and is not available at test time.

For both settings, we compare exemplars containing only uni-facet cases with those including both uni- and multi-facet cases. GPT-5 results are presented in \cref{tab:exact_match_ex}, with full A/U case-wise results reported in Appendix~\ref{appendix:results_appendix}, along with complementary statistical analyses.

\begin{table}[t]
\small
\centering
\begin{tabular}{lcccc}
\toprule
\textbf{Setting} & \multicolumn{2}{c}{Uni} & \multicolumn{2}{c}{Multi} \\
\cmidrule(lr){2-3} \cmidrule(lr){4-5}
 & LEM & SEM & LEM & SEM \\
\midrule
Zero-shot & 19.2 & 15.0 & 15.2 & 5.2 \\
No meta (uni) & 56.9 & 42.9 & 55.3 & 13.4 \\
No meta (uni+multi) & \textbf{62.7} & 50.3 & 61.4 & 23.4 \\
Meta (uni) & 60.5 & 53.9 & 67.0 & \textbf{60.2} \\
Meta (uni+multi) & 61.3 & \textbf{55.8} & \textbf{70.1} & 60.1 \\
\bottomrule
\end{tabular}
\caption{Few-shot single-turn SQL prediction for column ambiguity using GPT-5 (\%).}
\label{tab:exact_match_ex}
\end{table}

\begin{table*}[!t]
\small
\centering
\setlength{\tabcolsep}{4pt}
\renewcommand{\arraystretch}{1.08}
\begin{tabular}{lcccccccccccc}
\toprule
\textbf{Model} &
\multicolumn{8}{c}{\textbf{A/U Use Cases}} &
\multicolumn{4}{c}{\textbf{Conversation Types}} \\
\cmidrule(lr){2-9} \cmidrule(lr){10-13}
& \makecell{U-Lex\\Amb} &
  \makecell{M-Lex\\Amb} &
  \makecell{U-Sem\\Amb} &
  \makecell{M-Sem\\Amb} &
  \makecell{U-Col\\Unans} &
  \makecell{M-Col\\Unans} &
  \makecell{U-Val\\Amb} &
  \makecell{M-Val\\Amb} &
  Concise & Verbose & Partial & Not \\
\midrule
GPT-4o           & 74.2 & 73.2 & 74.2 & 67.2 & 99.0 & 100.0 & 68.3 & 60.6 & 73.1 & 80.5 & 74.6 & 62.6 \\
GPT-4.1          & 75.2 & 73.6 & 81.0 & 73.2 & 95.0 & 99.0  & 75.2 & 66.7 & 68.3 & 82.2 & 78.1 & 71.7 \\
GPT-5            & 73.0 & 71.6 & 77.6 & 69.4 & 83.0 & 93.0  & 78.3 & 70.8 & 62.9 & 79.3 & 77.2 & 70.0 \\
Grok-3 Mini Fast & 78.2 & 75.8 & 82.4 & 77.8 & 86.0 & 94.0  & 76.8 & 65.9 & 64.8 & 81.7 & 79.3 & 75.8 \\
LLaMA-3.3        & 76.8 & 76.8 & 80.6 & 69.8 & 44.0 & 72.0  & 72.6 & 64.9 & 42.3 & 81.3 & 76.5 & 72.8 \\
\bottomrule
\end{tabular}
\caption{Multi-turn SQL prediction performance on Spider, measured by execution accuracy (EA, \%). U = uni-facet; M = multi-facet; Lex = lexical; Sem = semantic; Col = column; Val = value; Unans = unanswerable.}
\label{tab:conv_results_spider}
\end{table*}

\begin{table}[t]
\centering
\small
\setlength{\tabcolsep}{4pt}
\begin{tabular}{llcccc}
\toprule
\textbf{Dataset} & \textbf{Method} &
\multicolumn{2}{c}{U Col Amb} &
\multicolumn{2}{c}{M Col Amb} \\
\cmidrule(lr){3-4} \cmidrule(lr){5-6}
& & DA & MA & DA & MA \\
\midrule
\multirow{2}{*}{Spider}
& AmbiSQL    & 99.6 & 57.8 & \textbf{100.0} & \textbf{20.4} \\
& AmbiSQL-CT & \textbf{99.8} & \textbf{62.2} & \textbf{100.0} & 19.6 \\
\midrule
\multirow{2}{*}{BIRD}
& AmbiSQL    & 98.9 & 50.1 & \textbf{100.0} & 5.4 \\
& AmbiSQL-CT & \textbf{99.3} & \textbf{57.7} & \textbf{100.0} & \textbf{7.0} \\
\bottomrule
\end{tabular}
\caption{Column ambiguity detection (\%).}
\label{tab:ad_gpt_col}
\end{table}

\begin{table}[t]
\centering
\small
\setlength{\tabcolsep}{4pt}
\begin{tabular}{llcccc}
\toprule
\textbf{Dataset} & \textbf{Method} &
\multicolumn{2}{c}{U Val Amb} &
\multicolumn{2}{c}{M Val Amb} \\
\cmidrule(lr){3-4} \cmidrule(lr){5-6}
& & DA & MA & DA & MA \\
\midrule
\multirow{2}{*}{Spider}
& AmbiSQL    & 98.0 & \textbf{22.0} & 96.8 & 35.1 \\
& AmbiSQL-CT & \textbf{100.0} & 21.8 & \textbf{100.0} & \textbf{40.8} \\
\midrule
\multirow{2}{*}{BIRD}
& AmbiSQL    & \textbf{98.8} & 37.7 & 98.0 & 26.0 \\
& AmbiSQL-CT & \textbf{98.8} & \textbf{41.3} & \textbf{98.0} & \textbf{32.0} \\
\bottomrule
\end{tabular}
\caption{Value ambiguity detection (\%).}
\label{tab:ad_gpt_val}
\end{table}

NL2SQL exemplars substantially improve both LEM and SEM over zero-shot prompting. Incorporating both uni- and multi-facet exemplars further outperforms uni-only settings, with particularly strong gains on multi-facet cases. Metadata-enhanced exemplars provide additional improvements, especially in SEM for multi-facet scenarios. As SEM reflects recovery of the complete set of valid SQL interpretations for an A/U query, it serves as a more diagnostic metric than LEM. Overall, the inclusion of structured metadata significantly enhances performance on multi-facet A/U cases. Further gains may be achievable through optimized exemplar selection.

\cref{tab:conv_results_spider} and \cref{tab:conv_results_bird} report zero-shot performance on the multi-turn SQL prediction task. Unanswerable cases are generally easier than ambiguous ones, as pivot terms do not correspond to any schema element and correctly yield a \texttt{null} output, resulting in higher EA. In contrast, multi-facet ambiguities are consistently more challenging than uni-facet cases, as reflected by lower EA. Performance differences across LLMs are relatively modest, with no single model dominating across all A/U categories, suggesting that successful ambiguity resolution depends more on interaction structure than on model-specific capabilities. EA also varies by conversation type: \textit{verbose} interactions achieve higher scores by providing stronger contextual clarification signals.

\cref{tab:ad_gpt_col,tab:ad_gpt_val}\footnote{Additional results are provided in Appendix~\ref{appendix:results_appendix}.} summarize AmbiSQL’s detection performance on the \textsc{Clarity} benchmark. While detection accuracy (DA) is near-perfect, match accuracy (MA) remains substantially lower, indicating weaknesses in schema-level localization. Errors primarily arise from misidentifying ambiguous terms and conflating lexical with semantic ambiguity. Because MA jointly evaluates detection and alignment with ground-truth metadata, models often correctly flag ambiguity but produce misaligned clarification outputs. Incorporating \textsc{Clarity} exemplars improves MA for uni-facet cases but yields limited gains for multi-facet settings, underscoring the importance of fine-grained, metadata-aware evaluation beyond binary detection.

We conduct paired $t$-tests to assess gains from \textsc{Clarity} exemplars. NL2SQL models augmented with \textsc{Clarity}-based few-shot examples achieve significant improvements over both zero-shot and standard few-shot prompting without metadata ($p < 0.005$). Similarly, AmbiSQL shows significant gains in column ambiguity detection with \textsc{Clarity} exemplars ($p < 0.005$).

\section{Related Work}

\textbf{Single-Turn Ambiguity and Unanswerability.}
Most benchmarks use rule-based or semi-automated pipelines combining LLMs and human annotation. Prior work generates ambiguity via schema augmentation or templates \citep{ambrosia,bhaskar2023benchmarking,wang2023know}, or synthesizes ambiguous NL2SQL examples from schema-level patterns \citep{squab}, while \citet{luo2025nvbench} study ambiguity in text-to-visualization. However, ambiguity is typically limited to a single failure mode per query with coarse instance-level labels.

\noindent\textbf{Conversational NL2SQL.}
 \citet{practiq} define a fixed clarification workflow via schema augmentation. \citet{mmsql} generate multi-turn QA sequences by prompting over existing datasets, without explicitly modeling ambiguity resolution or unanswerable cases. \citet{birdinteract} transform single-turn NL2SQL instances into dialogues via expert annotation. These datasets assume cooperative users who provide clean, sufficient clarifications and ignore variation in response quality.

\noindent\textbf{NL2SQL Dataset Generation Frameworks.}
The state-of-the-art frameworks \cite{duan2025dsqg, li2025omnisql, guo2025sqlforge, caferouglu2025sing, pourreza2024sql} typically assume a single well-defined SQL target per query and do not explicitly address multiple ambiguities, unanswerabilities, or interactive resolutions.

\section{Conclusion and Future Work}

\textsc{Clarity} provides a controlled framework for evaluating NL2SQL systems under multi-faceted ambiguity, unanswerability, and diverse conversational settings. Rather than replacing existing benchmarks, it complements them by exposing schema-localization and ambiguity-resolution failures not captured by instance-level evaluation. Results show substantial variation across A/U types, indicating that NL2SQL systems should be explicitly optimized for diverse failure modes—an important requirement for enterprise deployment.

Future work includes modeling more realistic user behaviors, such as topic shifts and user uncertainty; handling multiple, heterogeneous forms of ambiguity; and supporting more complex multi-turn interactions beyond the settings considered in MMSQL \cite{mmsql} and BIRD-INTERACT \cite{birdinteract}. Extending \textsc{Clarity} to queries that require external or implicit knowledge, as well as integrating it into downstream systems through specialized fine-tuning and retrieval-augmented in-context learning, are promising directions for improving robustness under ambiguity.

\bibliography{paper}

\appendix
\onecolumn

\begingroup
\raggedbottom
\setlength{\floatsep}{10pt}
\setlength{\textfloatsep}{10pt}
\setlength{\intextsep}{10pt}

\section{The \textsc{Clarity} Framework - Appendix}\label{appendix:method_appendix}
This appendix presents a taxonomy of A/U in \textsc{Clarity}. We define core A/U concepts (Table~\ref{definition_examples}), characterize user clarification behaviors in multi-turn interactions (Table~\ref{tab:conv_types}), provide representative conversational examples (Table~\ref{tab:examples}), and analyze model failure patterns across A/U categories (Table~\ref{tab:failure_cases}).




\begin{table}[H]

\centering
\small
\resizebox{\textwidth}{!}{
\begin{tabular}{p{3cm} p{6.5cm} p{6cm}}
\hline
\textbf{Concept} & \textbf{Definition} & \textbf{Examples}\\
\hline
\textbf{Ambiguity} &
A term can be linked to \textbf{AT LEAST TWO} schema entities
(can be column names or column values).

\textbf{Subcategory:}
\begin{itemize}[nosep]
  \item \textit{Lexical Ambiguity:} Two terms or columns are lexically similar if they share a subset of tokens and writing style
  \item \textit{Semantic Ambiguity:} Two terms or columns are semantically similar if they share same meaning or similar underlying concept
\end{itemize} 
& \textbf{Lexical Ambiguity:} The term \texttt{date} is lexically ambiguous, as it may refer to either \texttt{Production\_Date} or \texttt{Sale\_Date} column in the schema.
\smallskip

\textbf{Semantic Ambiguity:} The term \texttt{performance} is semantically ambiguous, as it may refer to either \texttt{sales achieved} or \texttt{projects completed} column in the schema.\\
\hline
\textbf{Unanswerable} &
A term that \textbf{CANNOT} be linked to any schema entity (column names or column values) & Refer to examples mention under \textbf{Column} and \textbf{Value Use Cases}
\\
\hline
\textbf{Column Use Case} &
The A/U use case is linked with the column of the schema 
& 
\textbf{Unanswerable Column:} The marked column does not exist in the schema 

\smallskip
``how many \texttt{tables} did we have?''
\\
\hline
\textbf{Value Use Case} &
The A/U use case is linked with the value of the schema column & 
\textbf{Unanswerable Value:} The marked value does not exist in the database 

\smallskip
``show records where location is \texttt{Pru}''\\
\hline
\textbf{Facet-based Use Case} &
\textbf{\textit{Uni-facet:}} The natural language query contains single A/U column or value 

\smallskip
\textbf{\textit{Multi-facet:}} The query contains two or more A/U columns or values& 

The marked term represents ambiguity case
\smallskip

\textbf{\textit{Uni-facet:}} ``show me \texttt{name} grouped by city''
\smallskip
\textbf{\textit{Multi-facet:}} ``List the \texttt{date} claim where the \texttt{amount} is less than 200''

\\
\hline

\textbf{Single Turn Instance } &
One ambiguous natural language query associated with one or more SQL queries. None if unanswerable and no SQL query exists for unanswerable cases & 
\textbf{User Query:} ``List the date and amount''

\textbf{SQL Queries:} 

\textit{``SELECT paymentDate, amount FROM Order;''}

\smallskip
\textit{``SELECT dispatchDate, amount FROM Order;''}
\\
\hline

\textbf{Multi Turn or Conversational Instance } &
A series of user and agent natural language queries associated with a SQL query, where the series of conversations represents the resolution of ambiguous or unanswerable use case. & 

\textbf{User Query:} ``List the date where amount is less than 100''

\smallskip
\textbf{Agent Response:} ``When you say \texttt{date}, do you mean \texttt{paymentDate} or \texttt{dispatchDate}?''

\smallskip
\textbf{User Query:} ``Use paymentDate''

\smallskip
\textbf{Agent Response:} \textit{``SELECT paymentDate FROM Order WHERE amount \textless 200;''} 

\\
\hline

\end{tabular}
}
\caption{Ambiguity and Unanswerable (A/U) use case definitions.}
\label{definition_examples}
\end{table}



\begin{table}[H]
\centering
\footnotesize
\begin{tabular}{p{2.5cm} p{5cm} p{6cm}}
\hline
\textbf{Type} & \textbf{Definition} & \textbf{Examples} \\
\hline

\textbf{Helpful} &
User provides a response that helps in resolving the ambiguity found in the NL query.
\par\smallskip
\textbf{Subcategories:}
\par
\textit{Concise:} User response provides a direct answer.
\par
\textit{Verbose:} User response explains additional details.
&
\multirow{3}{6.5cm}{
\textbf{User:} show me name grouped by city
\par\smallskip
\textbf{Agent:} When you say ``name'', do you mean ``Customer Name'' or ``Product Name''?
\par\smallskip
\underline{\textbf{Concise:}}
\textbf{User:} Product Name
\par\smallskip
\underline{\textbf{Verbose:}}
\textbf{User:} I mean product names grouped by cities.
\par\medskip
\underline{\textbf{Partially Helpful:}}
\par
\textbf{User:} prod name (or just ``prod'')
\par\medskip
\underline{\textbf{Not Helpful:}}
\par
\textbf{User:} i don't know

}
\\
\cline{1-2}

\textbf{Partially Helpful} &
The user response only partially matches a schema entity prompted by the agent
&
\\
\cline{1-2}

\textbf{Not Helpful} &
The user response does not provide direct resolution instructions
&
\\
\hline

\end{tabular}
\caption{Different types of multi-turn conversation. The \textit{Not Helpful} case covers user responses that are irrelevant or out of scope, such as gibberish input or references to non-existent or unrelated schema columns.}

\label{tab:conv_types}
\end{table}

\begin{table}[H]
\small
\centering
\resizebox{\textwidth}{!}{
\begin{tabular}{l cc cc cc cc cc}
\toprule
\textbf{Model} &
\multicolumn{2}{c}{Unqualified for A/U} &
\multicolumn{2}{c}{Non-existent Target Group} &
\multicolumn{2}{c}{Term Evaluator Failure} &
\multicolumn{2}{c}{Query Evaluator Failure} &
\multicolumn{2}{c}{Data Screening Failure} \\

\cmidrule(lr){2-3} \cmidrule(lr){4-5} \cmidrule(lr){6-7} \cmidrule(lr){8-9}  \cmidrule(lr){10-11}
 & Spider & BIRD & Spider & BIRD  & Spider & BIRD & Spider & BIRD & Spider & BIRD\\
\midrule
Uni Lex Amb & 9.53 & 2.41 & 33.84 & 17.54  &22.94  &29.53  &22.13 & 37.35  &3.77 & 3.52 \\
Multi Lex Amb & 40.1 & 6.32  &32.01 & 49.48 & 15.73  &32.46 & 8.7 & 10.37  &0.03 & 0.26 \\
Uni Sem Amb & 9.53  & 2.41 & 20.38 & 13.95 & 36.47 & 35.79 & 24.48 & 28.68  &0.39 & 0.52\\
Multi Sem Amb & 40.1 & 6.32 & 26.09 & 28.55 & 25.61 &  52.61&  7.37 & 10.04  &0.01  &0.07 \\
Uni Val Amb &  73.06 & 71 & 0 & 0 & 12.2 & 11.56 & 5.4  &4.17 & 10.5 & 11.23 \\
Multi Val Amb &  99.8 & 96.31 & 0 & 0& 0 & 0.07 &0.04 & 4.46 & 0.03 & 0.45\\
Uni Col Unans & 9.53 & 2.41&  0 & 0 & 0.06 & 2.22 & 63.7 & 24.64 & 0.16 & 0.72\\
Multi Col Unans & 40.1 & 6.32 & 0 & 0 & 5.89  &17.21  &34.59 & 15.97 & 0 & 0\\
Uni Val Unans & 70.46 & 23.47 & 0  &0 & 0.06 & 0.02 & 5.11 & 2.42 & 0 & 0  \\
Multi Val Unans & 94.69 & 93.83 & 0 & 0.07 & 5.89 &  0.59 & 0.94 & 0.46 & 0& 0 \\
\bottomrule
\end{tabular}
}
\caption{Unsuccessful Cases (\%) at Various Steps of Framework . Total instances in original dataset Spider=7,000 and BIRD=1,534. Uni: Unifacet; Multi: Multifacet; Lex: Lexical Column; Sem: Semantic Column; Val: Value; Col: Column.}

\label{tab:failure_cases}
\end{table}

\begin{table}[H]

\centering
\footnotesize
\begin{tabular}{p{1cm} p{6cm} p{7cm}}
\toprule
\textbf{Example} & \textbf{Rejection Category  }                    & \textbf{Details  }     \\
\midrule
1       & Rejected due to implicit disambiguation & \textbf{Ambiguous Query:} What is the average number of horses for farms with total horses greater than 5000?
\par\smallskip
\textbf{Pivot Term: }horses
\par\smallskip
\textbf{Target Group: }{[}Working\_Horses, Total\_Horses{]}
\par\smallskip
\textbf{Reason for rejection:} The query context implicitly resolves the ambiguity (average → Working\_Horses, total → Total\_Horses), making the query effectively unambiguous  \par\smallskip    \\
\midrule
2       & Rejected due to weak ambiguity signal   &\textbf{ Ambiguous Query:}  What is the average number of horses for farms with total horses greater than 5000?
\par\smallskip
\textbf{Pivot Term:} "horses"
\par\smallskip 
\textbf{Target Group:} {[}Working\_Horses, Total\_Horses{]}
\par\smallskip
\textbf{Reason for rejection:} The query context implicitly resolves the ambiguity (average → Working\_Horses, total → Total\_Horses), making the query effectively unambiguous.\\
\bottomrule
\end{tabular}
\caption{Rejected instances identified by the NL Query Evaluator in the \textsc{CLARITY} framework.}

\end{table}



\begin{table}[H]
\centering
\footnotesize
\setlength{\tabcolsep}{6pt}
\renewcommand{\arraystretch}{1.2}
\begin{tabular}{%
  >{\raggedright\arraybackslash}p{2.25cm}
  >{\raggedright\arraybackslash}p{12.25cm}
}
\toprule
\textbf{A/U Use Case} & \textbf{Example} \\
\midrule

\textbf{Uni-facet Lexical Column Ambiguity Verbose Helpful} &

 "USER"  : "Find the list of attributes that appear more than 3 times."

\smallskip
 "AGENT"  : "Which attribute do you mean: attribute\_data\_type, attribute\_name, or attribute\_id?"

\smallskip
 "USER"  : "Please use attribute\_data\_type, that's what I meant."

\smallskip
 "AGENT"  : "SELECT attribute\_data\_type FROM Attribute\_Definitions GROUP BY attribute\_data\_type HAVING count(*)  >  3"

\\
\midrule

\textbf{Uni-facet Semantic Column Ambiguity Concise Helpful} &

 "USER"  : "Which region has the most wines with a score greater than 90?"

\smallskip
 "AGENT"  : "When you say "region," should I use County, Appelation, or state?"

\smallskip
 "USER"  : "County."

\smallskip
 "AGENT"  : "SELECT T1.County FROM APPELLATIONS AS T1 JOIN WINE AS T2 ON T1.Appelation  =  T2.Appelation WHERE T2.Score  >  90 GROUP BY T1.County ORDER BY count(*) DESC LIMIT 1"

\\
\midrule

\textbf{Multi-facet Lexical Column Ambiguity Partially Helpful} &
 "USER"  : "List the date claim for claims where the amount is less than or equal to the average amount."

\smallskip
 "AGENT"  : "When you say "date claim," should I use "date\_claim\_settled" or "Date\_Claim\_Made"? Also, for "amount," do you mean "Amount\_Settled" or "amount\_claimed"?"

\smallskip
 "USER"  : "made and settled"

\smallskip
 "AGENT"  : "SELECT Date\_Claim\_Made FROM Claims WHERE Amount\_Settled  <=  ( SELECT avg(Amount\_Settled) FROM Claims )"
\\
\midrule

\textbf{Multi-facet Lexical Column Ambiguity Not Helpful} &
 "USER"  : "What is the total lesson time where appellation is Janessa and identity component is Sawayn?"

\smallskip
 "AGENT"  : "When you say "appellation," are you referring to "first\_name", "middle\_name", or "last\_name"? Also, for "identity component," should that be "first\_name", "middle\_name", or "last\_name"?,"

\smallskip
 "USER"  : "I don't know."

\smallskip
 "AGENT"  : "The query cannot be generated as "appellation" and "identity component" cannot be mapped to valid column names"
\\
\midrule

\textbf{Uni-facet Value Ambiguity Concise Helpful} &
 "USER"  : "How many addresses are there in state "U.S. states"?"

\smallskip
 "AGENT"  : "When you say "U.S. states," which value in state\_province\_county do you mean? For example: Vermont, Georgia, Colorado, NorthCarolina, Arkansas, Pennsylvania, Kentucky, Michigan, Montana, NewYork, Missouri, Louisiana, or NewHampshire."

\smallskip
 "USER"  : "state\_province\_county = Colorado"

\smallskip
 "AGENT"  : "SELECT count(*) FROM addresses WHERE state\_province\_county  =  "Colorado""
\\
\midrule

\textbf{Uni-facet Column Unanswerable} &
 "USER"  : "Which customers have made at least two orders? Return the churn\_risk\_score and number of orders."

\smallskip
 "AGENT"  : "Sorry, I can't complete this request because the column churn\_risk\_score isn't found in the database, and there are no close matches."
\\
\midrule

\textbf{Uni-facet Value Unanswerable} &
 "USER"  : "Count the number of cards where card type code is Prepaid."

\smallskip
 "AGENT"  : "Sorry, I can't complete that request because the value 'Prepaid' does not exist in the 'card\_type\_code' column in the database."
\\

\bottomrule
\end{tabular}
\caption{Example conversations generated by \textsc{Clarity}.}
\label{tab:examples}
\end{table}

\endgroup


\clearpage

\section{Pseudo Codes - Appendix}\label{appendix:pseudocode}

In this section, we present the algorithms of each component in \textsc{CLARITY} following execution order. We start with the end-to-end framework and then elaborate each sub-module it invokes. 

\paragraph{Overview.}
Algorithm~\ref{alg:au-framework} summarizes the full \textsc{CLARITY} pipeline.
Given a database schema, an original NL--SQL pair, and a target ambiguity or unanswerability mode, the framework iteratively constructs pivot terms, generates ambiguous or unanswerable NL queries, simulates multi-turn conversations, and finally validates the resulting instances via automated screening.

\paragraph{Target Extraction.}
The first step identifies schema elements to be perturbed. Algorithm~\ref{alg:sqlparser} parses the SQL query and samples $K$ target columns or column--value pairs depending on the ambiguity or unanswerability mode.

\begin{algorithm}[htbp]
\caption{SQLParser}
\footnotesize
\label{alg:sqlparser}
\begin{algorithmic}
\REQUIRE SQL query $q$; integer $K$; \\ mode $m \in 
{\texttt{col\_amb}, 
\texttt{val\_amb}, 
\texttt{col\_unans}, 
\texttt{val\_unans}}$ 
\ENSURE Parsed representation $sql\_structure$; list $target$ of length $K$
\STATE $sql_structure \gets \textsc{ParseSQL}(q)$
\STATE $Columns \gets \textsc{ExtractColumns}(sql_structure)$
\STATE $ColValPairs \gets \textsc{ExtractColumnValuePairs}(sql_structure)$
\STATE $target \gets [\ ]$
\FOR{$k \gets 1$ to $K$}
    \IF{$m \in {\texttt{col\_amb}, \texttt{col\_unans}}$}
        \STATE $item \gets \textsc{RandomChoice}(Columns)$
        \STATE $target[k] \gets item$
        \STATE $Columns \gets Columns \setminus {item}$
    \ELSIF{$m \in {\texttt{val\_amb}, \texttt{val\_unans}}$}
        \STATE $item \gets \textsc{RandomChoice}(ColValPairs)$
        \STATE $target[k] \gets item$
        \STATE $ColValPairs \gets ColValPairs \setminus {item}$
    \ELSE
        \STATE \textbf{error} Unknown mode $m$
    \ENDIF
\ENDFOR \\
\STATE \textbf{return} $(target, sql\_structure)$
\end{algorithmic}

\end{algorithm}

\paragraph{Schema Retrieval.}
Given a target element, Algorithm~\ref{alg:adaptive-schema-retriever} retrieves a set of competing columns or values that serve as the basis for ambiguity or unanswerability construction. 
For ambiguous queries, retrieval is selective; for unanswerable queries, the full schema is used as a negative reference.

\begin{algorithm}[!htbp]
\caption{AdaptiveSchemaRetriever}
\footnotesize
\label{alg:adaptive-schema-retriever}
\begin{algorithmic}
\REQUIRE target $t$; \\ mode $m \in 
{\texttt{col\_amb}, 
\texttt{val\_amb}, 
\texttt{col\_unans}, 
\texttt{val\_unans}}$ ; \\
schema $\mathcal{S}$; $\textit{retrieval\_type} \in {\texttt{llm}, \texttt{hybrid}, \texttt{combined}}$
\ENSURE list $G$ of similar columns or values for A/U query generation
\STATE $sample_k \gets \textsc{HeuristicTopK}(|\mathcal{S}|)$
\IF{$m \in {\texttt{col\_amb}, \texttt{val\_amb}}$}
    \IF{$m = \texttt{col\_amb}$}
        \STATE $target_space \gets \mathcal{S}$ \COMMENT{schema column space}
    \ELSE
        \STATE $target_space \gets \textsc{AllValuesInColumn}(t,, sample_k)$
    \ENDIF
    \IF{$\textit{retrieval\_type} = \texttt{llm}$}
        \STATE $G \gets \textsc{LLMRetriever}(t,, target_space)$
    \ELSIF{$\textit{retrieval\_type} = \texttt{hybrid}$}
        \STATE $G \gets \textsc{DenseRetriever}(t,, target_space)$
    \ELSIF{$\textit{retrieval\_type} = \texttt{combined}$}
        \STATE $k \gets \textsc{HeuristicTopK}(|target_space|)$
        \STATE $candidates \gets \textsc{HybridRetriever\_TopK}(t,, target_space,, k)$
        \STATE $candidates_normalized \gets \textsc{NormalizeSimilarityScores}(candidates)$
        \STATE $G \gets \textsc{LLM\_Filter}(t,, candidates_normalized,, m)$
    \ELSE{}
        \STATE \textbf{error} Unknown retrieval type
    \ENDIF
\ELSIF{$m \in {\texttt{col\_unans}, \texttt{val\_unans}}$}
    \STATE $G \gets \mathcal{S}$ \COMMENT{use full schema as negative reference}
\ELSE
    \STATE \textbf{error} Unknown mode $m$
\ENDIF
\STATE \textbf{return} $G$
\end{algorithmic}
\end{algorithm}

\paragraph{Pivot Term Generation and Evaluation.}
Algorithms~\ref{alg:pivot-term-generator} and~\ref{alg:pivot-term-evaluator} jointly construct and validate synthetic pivot terms. 
The generator proposes candidate pivot terms, while the evaluator checks validity, dissimilarity from schema elements, and ambiguity consistency. A feedback loop ensures only high-quality pivot terms are retained.

\begin{algorithm}[!htbp]
\caption{PivotTermGenerator}
\footnotesize
\label{alg:pivot-term-generator}
\begin{algorithmic}
\REQUIRE $G$  set of retrieved columns or values; \\mode $m \in {\texttt{col\_amb}, 
\texttt{val\_amb}, 
\texttt{col\_unans}, 
\texttt{val\_unans}}$; 
$generation\_type \in {\texttt{semantic}, \texttt{lexical}}$
\ENSURE synthetic pivot term $p$ that induces ambiguity or unanswerability
\IF{$\textsc{Contains}(m, \texttt{ambiguity})$}

    \STATE $constraint \gets \textsc{ExtractDefinition}(generation\_type)$
\ELSE
    \STATE $constraint \gets \textsc{ExtractDefinition}(m)$
\ENDIF
\STATE $p \gets \textsc{LLM\_GeneratePivotTerm}(G, m, generation\_type, constraint)$
\STATE \textit{return} $p$
\end{algorithmic}
\end{algorithm}

\begin{algorithm}[!htbp]
\caption{PivotTermEvaluator}
\footnotesize
\label{alg:pivot-term-evaluator}
\begin{algorithmic}
\REQUIRE $p$ pivot term ; \\ $G$ retrieved set ; \\ $S$ database schema ; \\
mode $m \in {\texttt{col\_amb}, 
\texttt{val\_amb}, 
\texttt{col\_unans}, 
\texttt{val\_unans}}$; \\
$generation\_type \in {\texttt{semantic}, \texttt{lexical}}$;
\ENSURE $\texttt{evaluation\_outcome}$: dictionary of evaluation outcomes and feedbacks; $\texttt{all\_pass}$: boolean

    \STATE $\texttt{evaluation\_outcome} \gets$ empty dictionary
    \STATE $\texttt{evaluation\_outcome}[\texttt{``pivot\_validity''}] \gets \texttt{LLM}(p, m, \texttt{generation\_type})$ \COMMENT{Check $p$ satisfies the definition(s) in $m$}
    \STATE $\texttt{evaluation\_outcome}[\texttt{``pivot\_dissimilarity\_all''}] \gets \texttt{LLM}(p, S, m, \texttt{generation\_type})$ \COMMENT{Verify $p$ is dissimilar to schema $S$}
    \IF{$\textsc{Contains}(m, \texttt{ambiguity})$}
        \STATE $\texttt{evaluation\_outcome}[\texttt{pivot\_term\_for\_target\_group}]
        \gets \texttt{LLM}(p, G, m, \texttt{generation\_type})$
        \COMMENT{Check whether $p$ is truly ambiguous and conflicts with multiple schema columns or values in $G$}
    \ENDIF
    \STATE $\texttt{all\_pass} \gets \textsc{AllTrue}(\textsc{values}(\texttt{evaluation\_outcome}))$

    \STATE \textit{return} $\texttt{evaluation\_outcome}, \texttt{all\_pass}$

\end{algorithmic}
\end{algorithm}

\paragraph{NL Query Generation and Evaluation.}
Algorithm~\ref{alg:nlquerygenerator} injects validated pivot terms into the original NL query while preserving its intent and surface form. Algorithm~\ref{alg:nlqueryevaluator} then verifies pivot usage, ambiguity or unanswerability compliance, and semantic consistency with the original SQL query.

\begin{algorithm}[!htbp]
\caption{NLQueryGenerator}
\footnotesize
\label{alg:nlquerygenerator}
\begin{algorithmic}
\REQUIRE original NL query $u$; SQL query $q$; set of pivot terms $p$; set of targets $t$
\ENSURE A/U NL query $r_1$ containing all pivot terms
\STATE $\mathcal{C} \gets$ constraint set defined as:
\STATE \quad
(i) Use pivot terms in $p$ to refer to targets in $t$ (do not use specific target names).
\STATE \quad (ii) Ensure $r_1$ semantically corresponds to the SQL query $q$.
\STATE \quad (iii) Preserve the sentence structure, intent, tone, and writing style of $u$.
\STATE \quad (iv) Suppress contextual cues that would disambiguate targets in $t$ from pivot terms in $p$.
\STATE $r_1 \gets \textsc{LLM}\big(\textsc{Prompt}(u, q, p, t, \mathcal{C})\big)$
\STATE \textit{return} $r_1$
\end{algorithmic}
\end{algorithm}

\begin{algorithm}[!htbp]
\caption{NLQueryEvaluator}
\footnotesize
\label{alg:nlqueryevaluator}
\begin{algorithmic}
\REQUIRE $r_1$ generated A/U NL query; \\
 $q$ SQL query; \\ $p$ pivot terms; \\ $t$ targets; \\ $S$ schema; \\ mode $m \in {\texttt{col\_amb}, \texttt{val\_amb}, \texttt{col\_unans}, \texttt{val\_unans}}$; \\
 generation\_type $\in {\texttt{semantic}, \texttt{lexical}}$
\ENSURE evaluation\_outcome: dictionary mapping criteria to (pass, feedback); all\_pass: boolean
\STATE evaluation\_outcome $\gets$ empty dictionary
\STATE evaluation\_outcome[``pivot\_in\_utterance''] $\gets$ \textsc{LLM}$(r_1, p)$ \COMMENT{Verify all pivot terms in $p$ appear in $r_1$}
\STATE evaluation\_outcome[``validity''] $\gets$ \textsc{LLM}$(r_1, m, \textit{generation\_type}, S)$ \COMMENT{Check A/U compliance for $m$ and \textit{generation\_type}; ensure no contextual disambiguation}
\STATE evaluation\_outcome[``consistency''] $\gets$ \textsc{LLM}$(r_1, p, t, q)$ \COMMENT{Check semantic consistency with $q$ after resolving $p$ via $t$}
\STATE $\texttt{all\_pass} \gets \textsc{AllTrue}(\textsc{values}(\texttt{evaluation\_outcome}))$
\STATE \textit{return} (all\_pass, evaluation\_outcome)

\end{algorithmic}
\end{algorithm}

\paragraph{Conversation Modeling.}
Given a validated ambiguous or unanswerable NL query, Algorithm~\ref{alg:conversation-generator} simulates a multi-turn interaction between a user and an agent. Different response styles (helpful, partially helpful, or unhelpful) are supported to model realistic clarification behaviors.

\begin{algorithm}[!htbp]
\caption{ConversationGenerator}
\footnotesize
\label{alg:conversation-generator}
\begin{algorithmic}
\REQUIRE $r_1$ initial NL query; \\ $q$ original SQL query; \\ $p$ set of pivot terms; \\ $t$ set of targets; \\ $m mode  \in {\texttt{col\_amb}, \texttt{val\_amb}, \texttt{col\_unans}, \texttt{val\_unans}}$; \\ $R response\_type  \in {\texttt{concise helpful}, \texttt{verbose helpful}, \texttt{partial}, \texttt{unhelpful}}$
\ENSURE conversation log $C = {(r_t, t_t)}_{t=1}^{T}$ (list of user/agent response tuples)

\STATE $t_1 \gets \textsc{GenerateTemplateResponse}({r_1, p, t, m})$ \COMMENT{Template indicating A/U use case}
\STATE $C \gets \textsc{GenerateConversation}({t_1, R, q})$ \COMMENT{Generate multi-turn user/agent responses using LLM}
\STATE \textit{return} $C$

\end{algorithmic}
\end{algorithm}

\paragraph{Final Validation.}
Algorithm~\ref{alg:automated-data-screening} performs automated quality control using predefined test cases specific to each ambiguity or unanswerability mode. Only instances that pass all screening checks are retained in the final dataset.

\begin{algorithm}[!htbp]
\caption{AutomatedDataScreening}
\footnotesize
\label{alg:automated-data-screening}
\begin{algorithmic}
\REQUIRE  $C$ Multi--turn instance; $p$ set of pivot terms; $t$ set of targets; $S$ schema; mode $m \in {\texttt{``col\_amb''}, \texttt{``val\_amb''}, \texttt{``col\_unans''}, \texttt{``value\_unans''}}$
\ENSURE validation\_flag $\in {\text{True}, \text{False}}$
\STATE $TC \gets \textsc{ExtractPredefinedTestCase}(m)$
\STATE evaluation $\gets []$  \COMMENT{initialize empty evaluation}
\FOR{$i \gets 1$ to $|TC|$}
\STATE evaluation[$i$] $\gets \textsc{LLM}(TC[i], C, p, t, S)$
\ENDFOR
\STATE \textit{return} True if all items in evaluation pass, else False
\end{algorithmic}
\end{algorithm}

\clearpage

\subsection{LLM Prompts}
The prompts utilised in \textsc{Clarity} are presented below.

\begin{tcolorbox}[mymultipagebox]
\captionof*{prompt}{A/U Term Generator Prompt - Lexical Column Ambiguity Case}
You are a SQL expert specializing in natural language (NL)-to-SQL translation and data generation.\\
You are given a column of interest and list of similar columns. Your task is to analyze the list of similar columns and identify a subset of columns that similar to the given column of interest, which might cause ambiguity if referenced without enough detail in a query. Ambiguity arises when multiple columns have similar names or overlapping meanings, making it unclear which one should be used.  \\
\\
\#\#\# **Instructions**\\
The following instructions must be strictly followed:\\
1. **Group Selection**: Identify groups of columns that are lexically similar to the given column of interest  \\
   - Identify the columns that have similar names and writing style or structure to the column of interest and group them. \\
   - The column of interest and selected columns must share common tokens, words or segments.\\
     - If there are many column names which are similar to the column of interest, then select the ones that share most tokens, words or segments.\\
   - The group should contain one or more columns from the given list.\\

2. **Term Generation**: Generate an ambiguous term for the column of interest that can likely be asked by a human database manager and cause ambiguity with the selected columns group using the following rules:\\
   - The ambiguous term should be lexically similar to column of interest in terms of wording, length, and style.\\
   - The ambiguous term should NOT be an exact match with column of interest and any column from the selected column list i.e., the term should be slightly different to all the columns in the list and should not exactly match with any of the columns.\\
   - The ambiguous term should NOT contain token, word, or segment from column(s) of the selected group which does not appear in the column of interest e.g., for "employee name" column of interest, "manager employee" is an invalid as token \verb|'|manager\verb|'| token exists in selected column groups, ["Manager Employee Name", "Manager Employee ID", "Manager Employee Contact"] but not in the column of interest.\\
   - The ambiguous term should NOT contain keyword or token that could be linked with a unique column e.g., for column of interest "Customer Number" and selected columns ["Customer Returning ID", "Customer Name"], "customer identifier" is an invalid ambiguous term as it can be associated with "customer Returning ID" instead of the column of interest. Moreover, it is also not associated with the column of interest.\\
   - The ambiguous term should NOT be a synonym or paraphrased word e.g., if we have two columns "Employee Title" and "Employee Designation" then "worker" term is invalid as it is a synonym of "Employee".\\
   - Do not use vague generic terms. For example, "count" for \verb|'|Number of People\verb|'|, \verb|'|Number of Cities\verb|'| or \verb|'|Number of Countries\verb|'| columns is an invalid ambiguous term.     \\
   - Ignore the case (uppercase or lowercase), singular plural and minor variations when generating an ambiguous term for the selected group. For example:\\
     -\verb|'|Customer Count\verb|'|, \verb|'|customer count\verb|'|, \verb|'|CUSTOMER COUNT\verb|'|, \verb|'|customer\_count\verb|'| and \verb|'|customer counts\verb|'| represent the same column. \\
     - \verb|'|no.\verb|'|, \verb|'|number\verb|'| or \verb|'|\#\verb|'| represent the same word or token. \\
     - \verb|'|Customer Count\verb|'| and \verb|'|Count Customer\verb|'| represent the same column. \\
   - If an ambiguous term cannot be generated when multiple lexically similar columns do not exist, then use \verb|'|Null\verb|'| for ambiguous term.\\
3. **Constraints**:\\
   - The ambiguous term and selected group must be specific to the column of interest, i.e., the ambiguous term is generated for the column of interest which causes ambiguity with the selected columns group.\\
     - For the ambiguous term, DO NOT use tokens, words or segments from columns of the selected group that do not appear in the column of interest.\\
     - The token, words or segments in ambiguous term must appear in both column of interest and the columns of the selected group.\\
   - Take into consideration the history, if provided, when generating the response. \\
       - The history contain errors linked with the term generated in the previous attempt(s).\\
       - Strictly use the history to generate a valid ambiguous term.\\
   - Strictly generate the response considering the column of interest, similar columns list and given constraints. Do not introduce any assumptions or external knowledge beyond the given columns. \\
   - Refrain from providing elaborative reasoning and only provide a short summarised reason, following the provided examples and output template. \\
4. **Format**: Strictly generate the response using the given output format. DO NOT generate any other content or text.\\
\\
\#\#\# **Examples**\\
   \\
Example 1:\\
Column of Interest: "Monthly Orders"\\
Given List of Similar Columns: ["Total Orders", "Yearly Orders", "Order Dispatch Date" , "Order Invoice ID"]\\
History: None\\
Selected Group: ["Total Orders", "Yearly Orders"]\\
Ambiguous Term: "orders"\\
Reason: "The column of interest \verb|'|Monthly Orders\verb|'| share common token \verb|'|orders\verb|'| with all the columns in the selected list and \verb|'|orders\verb|'| itself is not a column name in the given list."   \\
\\
Example 2:\\
Column of Interest: "Customer\_ID"\\
Given List of Similar Columns: ["Employee\_ID", "Employee\_Name", "Department\_ID", "Department\_Name", "Organisation\_ID"]\\
History: None\\
Selected Group: ["Employee\_ID", "Department\_ID", "Organisation\_ID"]\\
Ambiguous Term: "ID"\\
Reason: "The column of interest \verb|'|Customer\_ID\verb|'| has \verb|'|ID\verb|'| which also appears in the three selected columns and \verb|'|ID\verb|'| itself is not a column name in the given list."   \\
\\
Example 3:\\
Column of Interest: "Order Generated Transfer Number"\\
Given List of Similar Columns: ["Order Generated Bar Number", "Parcel Generated Bar Number", "Order Generated Dispatch Number"]\\
History:  None\\
Selected Group: ["Order Generated Bar Number", "Order Generated Dispatch Number"]\\
Ambiguous Term: "order generated number"\\
Reason: "\verb|'|Order Generated\verb|'| and \verb|'|number\verb|'| in the column of interest are common across the three selected columns. So \verb|'|order generated number\verb|'| ambiguous term could be used as this is an exact column in the list"\\
\\
Example 4:\\
Column of Interest: "Cost"\\
Given List of Similar Columns: ["Annual Cost", "Order Date", "Order Cost", "Annual Profit", "Cost per Item"]\\
History: "\verb|'|cost\verb|'| itself is a column name in the given list. So this is not a valid ambiguous term"\\
Selected Group: ["Annual Cost", "Order Cost"]\\
Ambiguous Term: "cost info"\\
Reason: "cost info is ambiguous enough to cover column of interest and all the columns in the selected group list, avoiding exact match with any one of them"\\
\\
Example 5:\\
Column of Interest: "Invoice Allocated Serial Number"\\
Given List of Similar Columns: ["Inventory Allocated Batch Number", "Order Dispatch Batch Number", "Inventory Allocated Serial Number", "Invoice Allocated Batch Number"]\\
History: None\\
Selected Group: ["Inventory Allocated Batch Number", "Invoice Allocated Batch Number"]\\
Ambiguous Term: "allocated batch number"\\
Reason: "\verb|'|allocated number\verb|'| generalises to column of interest and selected columns, causing ambiguity in the selection of exact column name"\\
\\
Example 6:\\
Column of Interest: "Manager\_First\_Name"\\
Given List of Similar Columns: ["Manager\_Contact", "Manager\_Email", "Manager\_Designation", "Department"]\\
History: None\\
Selected Group: ["Manager\_Contact", "Manager\_Email", "Manager\_Designation"]\\
Ambiguous Term: "Name"\\
Reason: "The column of interest \verb|'|Manager\_First\_Name\verb|'| contains \verb|'|Manager\verb|'| which also appears in the three selected columns and \verb|'|Manager\verb|'| itself is not a column name in the given list."  \\
\\
Example 7:\\
Column of Interest: "Number of AC DC services"\\
Given List of Similar Columns: ["Number of AC services", "AC services requested", "Price per AC service", "Number of paid services", "Number of unpaid services"]\\
History: "\verb|'|no. of AC services\verb|'| is an invalid ambiguous term as it represents the actual column "Number of AC services" in the given list"\\
Selected Group: ["Number of AC services", AC services requested", "Price per AC service"]\\
Ambiguous Term: "AC services"\\
Reason: "\verb|'|AC services\verb|'| term can cause ambiguity for the column of interest, making it unclear which column is being referred to and generalise to all the columns in the given list"\\
\\
Example 8: \\
Column of Interest: "Department"\\
Given List of Similar Columns: ["Department", "Section", "Category"]\\
History: None\\
Selected Group: []\\
Ambiguous Term: \verb|'|Null\verb|'|\\
Reason: "There is no column lexically similar to Department, hence an ambiguous term cannot be created"\\
\\
Example 9:\\
Column of Interest: "managerName"\\
Given List of Similar Columns: ["employeeName", "departmentName", "contactNo", "head"]\\
History: None\\
Selected Group: ["employeeName", "departmentName"]\\
Ambiguous Term: "Name"\\
Reason: "The column of interest \verb|'|managerName\verb|'| contains \verb|'|Name\verb|'| which also appears in the two selected columns and \verb|'|Name\verb|'| itself is not a column name in the given list."  \\
\\
Example 10:\\
Column of Interest: "transaction\_number"\\
Given List of Similar Columns: ["transaction\_id", "customer\_id", "customer\_contact", "credit\_card\_transaction", "amount\_transaction"]\\
History: None\\
Selected Group: ["transaction\_id", "credit\_card\_transaction", "amount\_transaction"]\\
Ambiguous Term: "transaction"\\
Reason: "The column of interest \verb|'|transaction\_number\verb|'| contains \verb|'|transaction\verb|'| which also appears in the three selected columns and \verb|'|transaction\verb|'| itself is not a column name in the given list."  \\
\\
\#\#\# **Inputs**\\
\\
Column of Interest:\\
\{coi\}\\
\\
List of Similar Columns:\\
\{subset\_columns\}\\
\\
History:\\
\{history\_term\}\\
\\
\#\#\# **Outputs**\\
\\
**Your response must follow these instructions**:\\
\{format\_instructions\}\\
\end{tcolorbox}

\begin{tcolorbox}[mymultipagebox]
\captionof*{prompt}{A/U Term Evaluator Prompt - Lexical Column Ambiguity Case}
You are a database analyst specializing in extracting relevant columns from a database. You are given a term and a list of columns. Your task is to analyze the list of columns and the given term to determine whether the term is lexically similar to the columns and represents an ambiguous term, without being an exact match or a direct representation of any of them.\\
 \\
\#\#\# **Definition**\\
\\
**Ambiguity**: Ambiguity arises between a term and list of columns if it is not clear which column name is being referred by the term in the given list. A valid ambiguous term can be linked with more than one column in the given list.\\
 * The following conditions must be met for a valid ambiguous term:\\
     * The ambiguous term should be lexically similar to two or more columns in the list in terms of wording, length, and style.\\
         * Lexical similarity involves sharing same tokens, words or segments.\\
         * If a term can be linked to two or more columns of the given list then it qualifies for valid ambiguity.\\
     * The entire ambiguous term should be linked with two or more columns. If the entire term cannot be linked with the given columns, then it is invalid. For example:\\
         * \verb|'|primary contact\verb|'|, the entire term, can be linked to "primary contact name" and "primary contact address" raising ambiguity in column selection. \\
         * \verb|'|secondary email\verb|'| cannot be linked to "contact email" and "secondary address" as the entire term cannot be associated with the mentioned two columns. Such cases do not qualify for valid ambiguity.\\
     * The ambiguous term must not exactly match the name of any column in the provided list.\\
         * An exact match is only allowed if it refers to a single, specific column in the list. This represents invalid ambiguity.\\
         * If the term can be linked with more than one columns, then it holds valid ambiguity.\\
     * The ambiguous term should not be a generic synonym or paraphrased word e.g., if the list contains columns "Employee Title" and "Employee Designation" then "person" term is invalid as it is a synonym of "Employee" and there is no column that could be lexically linked with "person" term.\\
 * **Constraints**\\
     * Ignore the case (uppercase or lowercase), singular plural and minor variations during the comparison and analysis For example:\\
         *\verb|'|Customer Count\verb|'|, \verb|'|customer count\verb|'|, \verb|'|CUSTOMER COUNT\verb|'|, \verb|'|customer\_count\verb|'| and \verb|'|customer counts\verb|'| represent the same column. \\
         * \verb|'|no.\verb|'|, \verb|'|number\verb|'| or \verb|'|\#\verb|'| represent the same word or token. \\
         * \verb|'|Customer Count\verb|'| and \verb|'|Count Customer\verb|'| represent the same column. \\
 * **Examples fo Valid Ambiguity:** \\
      * "student" is a valid ambiguous term for ["Student ID", "Program Start Date", "Course ID" , "Program End Date", "Student First Name", "Student Last Name" ] as it can be linked to multiple columns "Student ID", "Student First Name", and "Student Last Name" columns, causing ambiguity in column selection.\\
      * "Sales - Contact Number" is a valid ambiguous term for ["Sales - Customer Home Number", "Sales - Customer Office Number", "Sales - Customer ID", "Purchases - Employee ID"] as it causes ambiguity in selecting column either "Sales - Customer Home Number" or "Sales - Customer Office Number".\\
      * "employee" is a valid ambiguous term for ["name\_employee", "ID\_employee", "contact\_employee"] as "employee" segment is applicable to all the columns in the list and does not have an exact match with any of the columns.\\
      * "dept" is a valid ambiguous term for ["DeptID", "HeadID", "DeptName","HeadName", "HeadEmail", "DepartEmail"] as it can be maped to columns "DeptID", "DeptName", and "DepartEmail", not clarifying which one is being referred to.\\
      * "cost info" is a valid ambiguous term for ["Cost", "Annual Cost", "Monthly Cost"] as it can be associated with all the columns, as \verb|'|info\verb|'| could be linked with any of the columns.\\
      * "student\_name" is a valid ambiguous term for ["student\_id", "student\_first\_name", "student\_last\_name", "course\_enrolled", "program\_enrolled"] as the entire term can be linked to "student\_first\_name" and "student\_last\_name" columns.\\
      * "given name" is a valid ambiguous term for ["customer id", "customer given name", "customer last name", "employee id", "employee given name", "employee last name" ,"employee contact"] as the term can be linked with "employee given name" and "customer given name" columns.\\
 * **Examples of Invalid Ambiguity:**      \\
      * "purchase date" is an invalid ambiguous term for ["Purchase Date", "Purchase Date Approved", "Purchase Date Initiated"] as it has an exact match with "Purchase Date" in the given list.\\
      * "manager employee" is an invalid ambiguous term for ["Manager Employee Name", "Manager Employee StartDate", "Employee Manager", "Manager Employee Contact"] as it can be mapped to "Employee Manager" column.\\
      * "finance dept" is an invalid ambiguous term for ["Finance Department", "Corporate Finance Department", "Finance Planning Department", "HR Department"] as "finance dept" is another variant of "Finance Department" in the list and does not represent ambiguity.\\
      * "code" is an invalid ambiguous term for ["zipcode", "street number", "unit", "state" ] as "code" can directly map to "zipcode". To qualify for ambiguity criterion, the term must be lexically similar to at least two columns.\\
      * "employee salaries" is an invalid ambiguous term for ["Employee ID", "Employee Name", "Employee Salary", "Employee Salary Level", "Employee Designation"] as it represents a variant of "Employee Salary" column. Therefore, it does not qualify for ambiguity criterion.\\
      * "customer\_number" is an invalid ambiguous term for ["customer\_id", "customer\_registration", "customer\_contact", "employee\_contact", "employee\_number"] as the entire term cannot be linked with any of the columns in the list . Therefore, it does not qualify for ambiguity criterion.\\
\\
 \#\#\# **Instructions**\\
\\
 1.  **Analyze**: \\
      * Compare the `Given Term` with each column of the `Given Column List` using the given `Ambiguity` definition.\\
      * Focus on the conditions of the ambiguity criterion to identify valid ambiguous term.\\
 2.  **Decide**: Outcome is `"valid"` if the term is ambiguous. Otherwise, its `"invalid"`.\\
 3.  **Explain**: Provide a brief reason for your decision.\\
 4.  **Constraints**:\\
      * Base your reason strictly on the provided definitions.\\
      * The reason should be specific to the given term and list of columns.\\
      * Do not use any external knowledge or make assumptions.\\
 5.  **Format**: Structure your entire response according to the `Output Format` instructions\\
\\
 \\
 \#\#\# **Inputs**\\
\\
 Given Term:\\
 \{term\}\\
 \\
 Given List of Columns:\\
 \{dataset\_columns\}\\
\\
 \#\#\# **Outputs**\\
 \\
 **Your response must follow these instructions**:\\
 \{format\_instructions\}\\
\end{tcolorbox}

\begin{tcolorbox}[mymultipagebox]
\captionof*{prompt}{NL Query Generator Prompt}
You are a SQL expert specializing in natural language (NL)-to-SQL translation and data generation.\\
Your task is to generate a natural language query from the given SQL query. This query is then used to generate its ambiguous version.\\
The ambiguous version represents a natural language query that cannot be converted into SQL statement as the specification of column is not clear\\
\\
\#\#\# **Instructions**\\
\\
1. **Natural Language Query Generation**:\\
  - Using the given SQL query, write a natural language query that would generate the given SQL query\\
  - Strictly use the column names mentioned in the SQL query and its context to generate the natural language query. Do not make any assumptions\\
  - All the column names mentioned in the SQL query must be explicitly mentioned in the natural language query.\\
  - A reference natural language query is provided. Use similar writing style and tone to generate the natural language query.\\
2. **Ambiguous Query Generation**:\\
  - You are given a list of column names and corresponding ambiguous terms. Use these to generate an ambiguous query where the given column is replaced with the ambiguous term.\\
     - strictly use the given column name to replace it with the corresponding ambiguous term for the ambiguous query.\\
  - Use the generated natural language query and the given list of ambiguous candidates to generate its ambiguous variant as a user request.\\
  - Ensure the replacements create reasonable uncertainty about column references while still preserving interpretability.\\
  - Retain key contextual elements so that the ambiguous question still guides toward the SQL query but introduces a genuine need for clarification.\\
  - Ensure that the ambiguous version is genuinely unclear about which column is being referenced. Despite the ambiguity, the query should still make logical sense in the context of the SQL statement.\\
  - Strictly base the modifications on the given details and constraints. Do not introduce any assumptions or external knowledge beyond the schema.\\
3. **Constraints**:\\
  - The generated natural language query and its ambiguous variant should correspond to the same underlying SQL query. The aim of the queries should not change.\\
  - The generated natural language query should be written in a human-like style and should not use the exact column names format found in the SQL query. Use the provided examples as a guide.\\
    - Use natural language variant of the column name e.g., for column name "customer\_id", use "customer id"\\
    - If possible, avoid converting a singular ambiguous term into its plural form when generating the natural language query.\\
  - Write concise queries e.g, for where condition "amount>=10", the natural language query could be "for amount greater than or equal to".\\
  - Ensure the natural language query accurately reflects the SQL query and corresponds to each distinct command it contains.\\
  - Do not convert "joins" in the SQL query to natural query instruction.\\
  - Do not mention the database or table name in the generated natural language query.\\
  - Consider the history if available when generating the response. \\
     - The history contains issues with the generated query that mismatches with the given SQL query. Use this history to rewrite the queries to fix the error(s).\\
  - Examples are provided as guide to generate accurate response and avoid any mistakes.\\
4. **Format**:\\
  - Strictly generate response requested in the output instructions. \\
  - DO NOT GENERATE any other supplementary explanation or description. Strictly generate output as mentioned in the output format\\
\\
 \#\#\# **Examples**\\
 \\
 Example 1: Given SQL Query: \verb|'|SELECT count("Monthly Orders"), "Employee ID" FROM my\_data GROUP BY "Employee ID" ORDER BY Employee Name ASC\verb|'|\\
            Given Column Names: ["Monthly Orders"]\\
            Given Ambiguous Terms: ["orders"]\\
            History: None\\
            Reference Natural Language Query: "What is the total number of yearly orders by client ID sorted by client name?"\\
            Generated NL Query: "What are the number of monthly orders by employee id sorted on employee name"?\\
            Generated NL Ambiguous Query: "What are the number of orders by employee id sorted on employee name"?\\
\\
Example 2: Given SQL Query: \verb|'|SELECT "Order Detail", "Start Date" FROM my\_data WHERE "Start Date" = "01-01-2010" AND "Employee ID" = 454 ORDER BY Department ASC\verb|'|\\
           Given Column Names: ["Start Date", "Employee ID"]\\
           Given Ambiguous Terms: ["date", "employee"]\\
           History: None\\
           Reference Natural Language Query: "Display the client information for client id 178 with a start date of 11-11-2000, sorted by country."\\
           Generated NL Question: "display order detail when start date is 01-01-2010 and employee id is 454 sorted on department"\\
           Generated NL Ambiguous Query: "display order detail when date is 01-01-2010 and employee is 454 sorted on department"\\
\\
Example 3: Given SQL Query: "SELECT Host\_Name FROM farm\_competition WHERE Theme !=  \verb|'|Fantasy\verb|'| and Total\_Participants<=50"\\
           Given Column Names: ["Host\_Name"]\\
           Given Ambiguous Terms: ["host"]\\
           History: None\\
           Reference Natural Language Query: "Return the names of clients for internet services where the company is not \verb|'|Newtelco\verb|'| and number of client are less than or equal to 1000."\\
           Generated NL Question: \verb|'|Return the host name of competitions when theme is not "Fantasy" and total participants are less than or equal to 50?\verb|'|\\
           Generated NL Ambiguous Query: \verb|'|Return the host of competitions when theme is not "Fantasy" and total participants are less than or equal to 50?\verb|'|\\
\\
Example 4: Given SQL Query: "SELECT T1.product\_name, T2.sales\_price FROM Products AS T1 JOIN Product\_Pricing AS T2 ON T1.product\_id = T2.product\_id"\\
           Given Column Names: ["product\_name", "sales\_price"]\\
           Given Ambiguous Terms: ["product", "revenue"]\\
           History: None\\
           Reference Natural Language Query: "Can you provide the sales prices for each product name?"\\
           Generated NL Question: "Can you show me the sales prices for all the product names?"\\
           Generated NL Ambiguous Query: "Can you show me the revenue for all the products?"\\
\\
Example 5: Given SQL Query: \verb|'|SELECT P.ProjectName FROM Project AS P JOIN WorksOn AS W ON P.ProjectID = W.ProjectID GROUP BY P.ProjectID ORDER BY COUNT(*) ASC LIMIT 1\verb|'|\\
           Given Column Names: ["ProjectName"]\\
           Given Ambiguous Terms: ["name"]\\
           History: None\\
           Reference Natural Language Query: "What project has the lowest number of employees?"\\
           Generated NL Question: "What is the project name with fewest employees?"\\
           Generated NL Ambiguous Query: "What is the project name with fewest employees?"   \\
\\
Example 6: Given SQL Query: "SELECT T1.customer\_name, T2.billing\_address, T2.order\_amount FROM Customers AS T1 JOIN Orders AS T2 ON T1.customer\_id = T2.customer\_id"\\
           Given Column Name: ["order\_amount", "billing\_address"]\\
           Given Ambiguous Term: ["expense", "address"]\\
           History: None\\
           Reference Natural Language Query: "What are the names of customers, their billing addresses, and the amounts of their orders?"\\
           Generated NL Question: "What are the customer names, billing addresses, and order amounts?"\\
           Generated NL Ambiguous Query: "What are the customer names, addresses, and expenses?"\\
\\
Example 7: Given SQL Query: "SELECT C.customer\_number, C.residential\_address FROM Customers AS C JOIN Service\_Requests AS S ON C.customer\_number = S.customer\_id GROUP BY C.customer\_number ORDER BY COUNT(*) DESC LIMIT 1"\\
           Given Column Names: ["residential\_address"]\\
           Given Ambiguous Terms: ["personal information"]\\
           History: None\\
           Reference Natural Language Query: "Which customer submitted the highest number of service requests? Display their customer ID and home address."\\
           Generated NL Question: "Which customer has submitted the most service requests? Return the customer number and residential addresses."\\
           Generated NL Ambiguous Query: "Which customer has submitted the most service requests? Return the customer number and personal information."\\
\\
Example 8: Given SQL Query: \verb|'|SELECT "Item Code", "Production Date" FROM my\_dataset GROUP BY "Item Code" ORDER BY count(*) DESC LIMIT 1\verb|'|\\
           Given Column Names: ["Item Code", "Production Date"]\\
           Given Ambiguous Terms: ["item", "date"]\\
           History: None\\
           Reference Natural Language Query: "What item code appears most frequently, and what is its production date?"\\
           Generated NL Question: "What is the most common item code and its production date?"\\
           Generated NL Ambiguous Query: "What is the most common item and its date?"\\
\\
Example 9: Given SQL Query: \verb|'|SELECT "Item Type",  avg("Cost") FROM my\_dataset GROUP BY "Item ID"\verb|'|\\
           Given Column Names: ["Item Type"]\\
           Given Ambiguous Terms: ["item"]\\
           History: None\\
           Reference Natural Language Query: "What is the average price for every type of item?"\\
           Generated NL Question: "What is the average cost for each item type?"\\
           Generated NL Ambiguous Query: "What is the average cost for each item?"    \\
\\
Example 10: Given SQL Query: \verb|'|SELECT count("Annual Orders"), "Vendor ID" FROM Sales where Year = 2000 GROUP BY "Vendor ID"\verb|'| \verb|'|\\
            Given Column Names: ["Annual Orders"]\\
            Given Ambiguous Terms: ["purchases"]\\
            History: None\\
            Reference Natural Language Query: "What is the total number of yearly orders by vendor id for year 2000?"\\
            Generated NL Query: "What are the number of annual orders by vendor id for year 2000"?\\
            Generated NL Ambiguous Query: "What are the number of purchases by vendor id for year 2000"?\\
\\
\\
\#\#\# **Inputs**\\
Given SQL Query:\\
\{expected\_query\}\\
\\
Given Column Names:\\
\{cois\}\\
\end{tcolorbox}

\begin{tcolorbox}[mymultipagebox]
\captionof*{prompt}{NL Query Evaluator Prompt - Consistency with SQL Query}
You are a SQL expert specializing in natural language (NL)-to-SQL translation and data quality refinement.\\
You are given an ambiguous version of a natural language query. This ambiguity represents the use of generic words/terminology for a column which could be mapped to multiple columns in the database. This confusion can lead to different SQl query translation. Your task is to assess the consistency of the ambiguous version of the given query against the original SQL query.\\
   \\
\#\#\# **Instructions**\\
\\
The following instructions must be strictly followed:\\
1. **Consistency Assessment**: The ambiguous query should be compared with the original SQL query for assessment.\\
      - The ambiguous term(s) and corresponding true column(s) is given that resolves the ambiguity in the given query.\\
      - When the ambiguous term(s) is replaced by the true column(s), the resulting query should be consistent with the underlying SQL query. Specifically, it should meet the following two conditions:\\
         * the query translates to the given SQL query.\\
         * the intention and goal of the given ambiguous query aligns with the SQL query. \\
2. **Response Generation**: After replacing the ambiguous term(s) with true column(s) in the given ambiguous query, if the resulting query corresponds to underlying SQL query then the outcome is \verb|'|valid\verb|'|. Otherwise, it is \verb|'|invalid\verb|'|.\\
      - The valid assessment must address the two consistency condition.\\
      - Provide a reason for the assessment as well.\\
3.  **Constraints**:\\
      - Base your reason strictly on the provided instructions. Response should be specific to given queries, ambiguous term and true column information.\\
      - Do not use any external knowledge or make assumptions.\\
      - Refrain from providing a SQL language-based response and instead, provide a natural language response.\\
4. **Response Format**: Strictly generate the response using the given output format. DO NOT generate any other content or text.\\
   \\
\#\#\# **Examples**\\
   \\
Example 1: \\
Ambiguous Natural Language Query: \verb|'|show count of orders by employee sorted on employee name\verb|'|\\
Ambiguous terms: [\verb|'|orders\verb|'|, \verb|'|employee\verb|'|]\\
True Columns: [\verb|'|Monthly Orders\verb|'|, \verb|'|Employee ID\verb|'|]\\
Underlying SQL Query: SELECT count("Monthly Orders"), "Employee ID" FROM my\_data GROUP BY "Employee ID" ORDER BY "Employee Name" ASC\\
Outcome: \verb|'|valid\verb|'|\\
Reason:  \verb|'|When "orders" and "employee" in the ambiguous query are replaced by given true columns "Monthly Orders" and "Employee ID", it results in the same expected SQL query. Also, the goal and intention of the query is consistent with the SQL query.\verb|'|\\
\\
Example 2: \\
Ambiguous Natural Language Query: \verb|'|display the total detail for employee id 454 and start date of 01-01-2010, sorted by Department."\\
Ambiguous Terms: [\verb|'|detail\verb|'|]\\
True Column: [\verb|'|Order Detail\verb|'|]\\
Underlying SQL Query: \verb|'|SELECT "Order Detail", "Start Date" FROM my\_data WHERE "Start Date" = "01-01-2010" AND "Employee ID" = 454 ORDER BY Department ASC\verb|'|\\
Outcome: \verb|'|invalid\verb|'|\\
Reason: \verb|'|When "detail" in the ambiguous query is replace by given true column "Order Details", it resulting query does not matches with the original query and results in a different SQL query to the given one. Specifically, the ambiguous query mentions "total details" that suggests aggregation, which is inconsistent with the underlying SQL query.\verb|'|\\
\\
Example 3:\\
Ambiguous Natural Language Query: \verb|'|Can you show me the prices for all the products?"\\
Ambiguous Terms: [\verb|'|product\verb|'|, \verb|'|price\verb|'|]\\
True Column: [\verb|'|product\_name\verb|'|, \verb|'|billing\_price\verb|'|]\\
Underlying SQL Query: \verb|'|SELECT T1.product\_name, T2.billing\_price FROM Products AS T1 JOIN Product\_Pricing AS T2 ON T1.product\_id = T2.product\_id\verb|'|\\
Outcome: \verb|'|valid\verb|'|\\
Reason: \verb|'|When "product" and "price" in the ambiguous query is replace by given true columns "product\_name" and "billing\_price", it results in the same expected SQL query. Also, the goal and intention of the query is consistent with the underlying SQL query".\\
\\
Example 4: \\
Ambiguous Natural Language Query: "Count the number of membership"\\
Ambiguous Terms: [\verb|'|membership\verb|'|]\\
True Column: [\verb|'|Enrolment ID\verb|'|]\\
Underlying SQL Query: \verb|'|SELECT count(DISTINCT "Enrolment ID") FROM my\_dataset\verb|'|\\
Outcome: \verb|'|invalid\verb|'|\\
Reason: \verb|'|The SQL query is requesting for distinct "Enrolment ID", whereas the the ambiguous query is not considering the distinct "membership" which is linked with "Enrolment ID". As the ambiguous query is not consistent with the underlying SQL query, so the outcome is invalid".\\
\\
\#\#\# **Inputs**\\
\\
Ambiguous Natural Language Query:\\
\{ac\_query\}\\
   \\
Ambiguous Terms:\\
\{term\}\\
\\
True Columns:\\
\{cois\}\\
\\
Underlying SQL Query:\\
\{expected\_query\}\\
\\
\\
\#\#\# **Outputs**\\
**Your response must follow these instructions**:\\
\{format\_instructions\}\\
\end{tcolorbox}

\begin{tcolorbox}[mymultipagebox]
\captionof*{prompt}{Data Screening Prompt A}
You are a NL2SQL system expert. You are given a natural language query that belongs to one of the four categories:\\
**Type of Categories*\\
1. **Lexical Column Ambiguity**: The query includes tokens or terms that refer to a column, but there is no exact match with any column in the given schema. However, the term or token is lexically similar to two or more columns of the schema, making it unclear which column the query is referring to.\\
   - Lexical similarity means that the term shares similar name and writing style or structure with two or more columns of the give schema. \\
   - This ambiguity requires clarification from human for column selection to successfully generate a SQL query. Without this clarification, a SQL query cannot be generated\\
   - Here are a few examples of lexical similarity:\\
         - "orders" term is lexically similar to "Total Orders" and "Yearly Orders" columns\\
         - "ID" term is lexically similar to "Employee\_ID", "Department\_ID", and  "Organisation\_ID"\\
         - "allocated batch number" term is lexically similar to "Inventory Allocated Batch Number" and "Invoice Allocated Batch Number"\\
2. **Semantic Column Ambiguity**: The query includes tokens or terms that refer to a column, but there is no exact match with any column in the schema. However, the term or token is semantically similar to two or more columns of the schema, making it is unclear which column the query is referring to.\\
   - Semantic Similarity means that the term represents higher-level category, concept, synonym, or overlapping ideas in a knowledge space with two or more columns of the schema. \\
   - This ambiguity requires clarification from human for column selection to successfully generate a SQL query. Without this clarification, a SQL query cannot be generated\\
   - Here are a few examples of semantic similarity:\\
         - "issue" is semantically similar to "Security Breach", "Malfunctions in Hardware", and  "Hardware Failure" columns\\
         - "expense" is semantically similar to "annual\_cost", "order\_cost", "registration\_fee", and "seller\_commission\_fee" columns.\\
         - "performance" is semantically similar to "salesAchieved", "projectsCompleted", and "scoreReceived" columns\\
3. **Column Confusion**: The query contains tokens or terms that cannot be mapped to any columns in the schema. The term or token is neither lexically nor semantically similar to any columns in the schema. The column does not exists in the schema.\\
4. **Unambiguous**: The query contains tokens or terms referring to a column, with each term being mapped to a single column in the schema. Unambiguous queries can be translated to SQL query without any human intervention\\
\\
\#\#\# **Instructions**:\\
You are given a natural language query that may belong to one of the given categories. Your task is to identify terms or tokens that represents a column name but does not have an exact match with columns of the given schema. Follow the given instructions to achieve this goal.\\
1. Analyse the natural language query to identify the terms that refers to column in the schema taking into account the given ***Type of Categories**\\
   - Schema can contain multiple tables, compare the terms against each table\\
   - Ignore the case (uppercase or lowercase), singular plural and minor variations during the comparison. For example:\\
         -\verb|'|Customer Count\verb|'|, \verb|'|customer count\verb|'|, \verb|'|CUSTOMER COUNT\verb|'|, \verb|'|customer\_count\verb|'| and \verb|'|customer counts\verb|'| represent the same entity. \\
         - \verb|'|no.\verb|'|, \verb|'|number\verb|'| or \verb|'|\#\verb|'| represent the same word or token. \\
         - \verb|'|Customer Count\verb|'| and \verb|'|Count Customer\verb|'| represent the same entity. \\
         - \verb|'|Identifier\verb|'|, \verb|'|ID\verb|'|, or \verb|'|identity\verb|'| represent the same entity.\\
2. Identify the list of terms that refer to a column but do not exactly match any column names in the schema.\\
   - There could be more than one terms whose exact match does not exist\\
   - If exact match with the schema columns exist for the identified terms then use \verb|'|None Found\verb|'| for response\\
3. Provide a reason for the assessment. The reason should be specific to the given schema.\\
\\
**Format**: Strictly generate the response using the given output format, recording terms and reason seperately. DO NOT generate any other content or text.\\
\\
\#\#\# **Inputs**\\
Datbase Schema:\\
\{db\_schema\}\\
\\
Natural Language Question:\\
\{ac\_query\}\\
\\
\#\#\# **Outputs**\\
\{format\_instructions\}\\
\end{tcolorbox}

\begin{tcolorbox}[mymultipagebox]
\captionof*{prompt}{Data Screening Prompt B}
You are a data scientist. You are given two set of list, your task it is assess whether there is an overlap of the elements in the two lists or not.\\
\#\#\# **Instructions**:\\
- Compare the elements of given `List 1` and `List 2`.\\
- If there is an overlap of elements between the two lists i.e., the lists share common elements, then the response is \verb|'|yes\verb|'|, otherwise, its \verb|'|no\verb|'|.\\
   - The response is \verb|'|no\verb|'| if one of the lists does not contain valid elements for comparison\\
- Provide a reason for the outcome stating the overlapping or unique elements found during the comparison\\
- When making the comparison between the two lists, ignore the case (uppercase or lowercase), singular plural and minor variations . For example:\\
   -\verb|'|Customer Count\verb|'|, \verb|'|customer count\verb|'|, \verb|'|CUSTOMER COUNT\verb|'|, \verb|'|customer\_count\verb|'| and \verb|'|customer counts\verb|'| represent the same entity. \\
   - \verb|'|no.\verb|'|, \verb|'|number\verb|'| or \verb|'|\#\verb|'| represent the same word or token. \\
   - \verb|'|customer\_count\verb|'| and \verb|'|Count Customer\verb|'| represent the same entity. \\
- Strictly generate the response using the given output format. DO NOT generate any other content or text.\\
\\
List 1: \{term\}\\
List 2: \{cols\_to\_compare\}\\
\\
\#\#\# **Outputs**\\
\{format\_instructions\}\\
\end{tcolorbox}

\clearpage

\section{Evaluation Tasks - Appendix} \label{appendix:eval_appendix}

\begin{tcolorbox}[fontupper=\footnotesize, mymultipagebox]
\captionof*{prompt}{No-shot prompt for text-to-SQL queries generation of column ambiguity use case}

 The task is to write SQL queries based on the provided questions in English. Questions can take the form of an instruction or command and can be ambiguous, meaning they can be interpreted in different ways. 
        In such cases, write all possible SQL queries corresponding to different interpretations and separate each SQL query with ';' and an empty line. Do not include any explanations, and do not select extra columns beyond those requested in the question.\\
        
        Given the following Database schema:\\
        \{schema\}\\
        Answer the following:\\
        \{question\}
\end{tcolorbox}

\begin{tcolorbox}[fontupper=\footnotesize, mymultipagebox]
\captionof*{prompt}{No-shot prompt for conversation-to-SQL query generation}
You are an NL2SQL expert and you are given a user and agent conversation. This converstaion represent an initial user natural language query that can be ambiguous, which is later resolved by agent and user interaction.\\

        Use this conversation to generate a SQL query. Consider the given database schema to generate a SQL query. If a SQL query cannot be generated then generate "Null" response
        Do not include any explanations, and do not select extra columns beyond those requested in the question.\\

        Given the following Database schema:\\
        \{schema\}\\
        Given User-Agent Conversation:\\
        \{question\}

\end{tcolorbox}

\begin{tcolorbox}[fontupper=\footnotesize, mymultipagebox]
\captionof*{prompt}{Prompt for comparing the AmbiSQL results with the ground-truth metadata}

\#\#\# Task\\

    You are provided with a natural language question that contains ambiguity, making it impossible to directly convert it into a valid SQL query. A separate model detects this ambiguity and generates a clarification question along with a list of suggested columns that could be used to form a valid SQL query.\\
    
    Your task is to evaluate the clarification question and suggested columns by comparing them with the ground truth:\\
         - **Underlying Ambiguous Term**: The term in the natural language question that introduces ambiguity.\\
        - **Ground-Truth Columns**: The set of columns that could be used for the underlying ambiguous term to generate a valid SQL query.\\
    - After comparison, make a binary decision:\\
        - True: If the clarification question and suggested columns appropriately address the ambiguity and align with the ground truth.\\
        - False: If they do not properly address the ambiguity or correspond to the correct columns.
        - Provide a brief explanation justifying your decision\\

    \#\#\# Inputs\\
    Question: \{question\}\\

    Generated Clarification Question: \{clarification\}\\

    Suggested Columns: \{suggested\_columns\}\\

    Underlying Ambiguous term: \{term\}\\

    Ground-Truth Columns: \{gt\_columns\}\\
\end{tcolorbox}

\clearpage

\section{Results - Appendix} \label{appendix:results_appendix}


\begin{table*}[!h]
\small
\centering
\resizebox{\textwidth}{!}{
\begin{tabular}{l cc cc cc cc}
\toprule
\textbf{Prompting Technique} &
\multicolumn{2}{c}{Unifacet Lexical Amb} &
\multicolumn{2}{c}{Multifacet Lexical Amb} &
\multicolumn{2}{c}{Unifacet Semantic Amb} &
\multicolumn{2}{c}{Multifacet Semantic Amb} \\

\cmidrule(lr){2-3} \cmidrule(lr){4-5} \cmidrule(lr){6-7} \cmidrule(lr){8-9} 
 & LEM & SEM & LEM & SEM  & LEM & SEM & LEM & SEM\\
\midrule
No-shot & 18.2 & 15.3 & 18 & 5.1 & 20.6 & 14.7 & 12.3 & 5.3 \\
  without metadata uni-facet & 55.9 & 42.4 & 61.5 &12.82 &57.8 & 43.3 &49.1 & 14.03\\
 without metadata uni-\& multi-facet  & \textbf{60.9} & 50.4 & 66.7 & 20.5 & \textbf{64.6} & 50.2 & 56.1 &26.3\\
with metadata uni-facet & 60.4 & 53.9 & 74.4 & 61.5 & 60.3 & 53.9 & 59.6 & \textbf{56.1}\\
with metadata uni-\& multi-facet & 58.6 & \textbf{54.5} & \textbf{76.9} & \textbf{64.1}  & 63.9 & \textbf{57} & \textbf{63.2} & \textbf{56.1}\\
\bottomrule
\end{tabular}
}
\caption{Few-Shot Single Turn SQL Prediction Task for Column Ambiguity (Amb) Use Cases Using GPT-5 (\%). Boldface indicates the comparatively best-performing value.}
\label{tab:exact_match_ex_detail}
\end{table*}

\begin{table*}[!h]
\small
\centering

\begin{tabular}{l cc cc cc cc}
\toprule
\textbf{\makecell{Prompting with \\ Exemplars}} &
\multicolumn{2}{c}{Unifacet Amb} &
\multicolumn{2}{c}{Multifacet Amb} \\
\cmidrule(lr){2-3} \cmidrule(lr){4-5} & LEM & SEM & LEM & SEM \\
\midrule
Zero-shot &  (17.3, 21.9) & (13, 17.1) & (8.6, 22.6) & (1.1, 10.7) \\
without metadata uni-facet  & (54.2, 60.2) & (40.1, 46) & (46.2, 66.7) & (7.5, 21.5)\\
without metadata uni-\& multi-facet  & (60.9, 66.6) & (47.4, 53.4) & (52.7, 72) & (16.1, 33.3)\\
with metadata uni-facet & (57.6, 63.4) & (51.2, 57) & (58.1, 77.4) & (50.5, 69.9)\\
with metadata uni-\& multi-facet & (58.7, 64.7) & (53.2, 59) & (61.3, 79.5) & (51.6, 71) \\
\bottomrule
\end{tabular}

\caption{Confidence intervals for the exact-match accuracy of the few-shot, single-turn SQL prediction task under column ambiguity (Amb) are reported in \ref{tab:exact_match_ex}. Confidence intervals are presented as (lower limit, upper limit) and are computed via bootstrapping with replacement over 10,000 resamples.}

\end{table*}


\begin{table*}[!h]
\small
\centering
\resizebox{\textwidth}{!}{
\begin{tabular}{l cc cc cc cc}
\toprule
\textbf{Model} &
\multicolumn{2}{c}{Unifacet Lexical Amb} &
\multicolumn{2}{c}{Multifacet Lexical Amb} &
\multicolumn{2}{c}{Unifacet Semantic Amb} &
\multicolumn{2}{c}{Multifacet Semantic Amb} \\

\cmidrule(lr){2-3} \cmidrule(lr){4-5} \cmidrule(lr){6-7} \cmidrule(lr){8-9} 
 & LEM & SEM & LEM & SEM  & LEM & SEM & LEM & SEM\\
\midrule
GPT-4o & 13.9 & 12.5 & 10.3 & 2.6 & 15.7 & 11.6 & 14 & 3.5\\
GPT-4.1 & 35.1  & 24.3 & 41  & 2.6 & 36.1  & 23.4  & 35.1  & 5.3\\
GPT-5 & 18.2 & 15.3 & 18 & 5.1 & 20.6 & 14.7 & 12.3 & 5.3\\
Grok-3 Mini Fast & 33.3 & 22.9 & 48.7 & 5.1 & 34.5 & 24.4& 35.1 & 8.8\\
LLaMA-3.3 & 31.2 & 21.8 & 18 & 2.6 & 33.8 & 25.2 &26.3 & 8.8\\
\bottomrule
\end{tabular}
}
\caption{No-Shot Single Turn SQL Prediction Task for Column Ambiguity (Amb) Use Cases (\%)}
\label{tab:exact_match}
\end{table*}

\begin{table*}[!h]

\small

\centering
\resizebox{\textwidth}{!}{
\begin{tabular}{l cc cc cc cc}
\toprule
\textbf{Prompting Technique} &
\multicolumn{2}{c}{Unifacet Lexical Amb} &
\multicolumn{2}{c}{Multifacet Lexical Amb} &
\multicolumn{2}{c}{Unifacet Semantic Amb} &
\multicolumn{2}{c}{Multifacet Semantic Amb} \\

\cmidrule(lr){2-3} \cmidrule(lr){4-5} \cmidrule(lr){6-7} \cmidrule(lr){8-9} 
 & LEM & SEM & LEM & SEM  & LEM & SEM & LEM & SEM\\
\midrule
Zero-shot & 13.9 & 12.5 & 10.3 & 2.6 & 15.7 & 11.6 & 14 & 3.5\\
  without metadata uni-facet & 25.5 & 19.8  &25.6 & 15.4 & 33  &25 & 31.6 & 7\\
 without metadata uni-\& multi-facet  & 26.3 & 21 & 23 & 12.8 & 33.3 & 26.8 & \textbf{40.4} & 10.5  \\
with metadata uni-facet & 29.2 & 24.3 & 17.9 & 7.7 & 39.7 & 30.3 & 15.8 & 12.3\\
with metadata uni-\& multi-facet & \textbf{38.6} & \textbf{35.7} & \textbf{35.9 }& \textbf{33.3} & \textbf{47.9} & \textbf{44.8} & 22.8 & \textbf{17.5}\\
\bottomrule
\end{tabular}
}
\caption{Few-Shot Single Turn SQL Prediction Task for Column Ambiguity (Amb) Use Cases Using GPT--4o(\%). Boldface indicates the comparatively best-performing value}
\end{table*}

\begin{table*}[!h]

\small

\centering
\resizebox{\textwidth}{!}{
\begin{tabular}{l cc cc cc cc}
\toprule
\textbf{Prompting Technique} &
\multicolumn{2}{c}{Unifacet Lexical Amb} &
\multicolumn{2}{c}{Multifacet Lexical Amb} &
\multicolumn{2}{c}{Unifacet Semantic Amb} &
\multicolumn{2}{c}{Multifacet Semantic Amb} \\

\cmidrule(lr){2-3} \cmidrule(lr){4-5} \cmidrule(lr){6-7} \cmidrule(lr){8-9} 
 & LEM & SEM & LEM & SEM  & LEM & SEM & LEM & SEM\\
\midrule
Zero-shot & 35.1  & 24.3 & 41  & 2.6 & 36.1  & 23.4  & 35.1  & 5.3\\
  without metadata uni-facet & 39.2  &28.2 & 41 & 10.3 & 49.8 & 36.8 & 40.4 & 14\\
 without metadata uni-\& multi-facet  & 39 & 29 & 43.6 & 10.3 & 49.8 & 38.6  &40.4 & 10.5 \\
with metadata uni-facet & 57.1 & 52.9 & 69.2 & 53.8 & 64.2 & 59 & 50.9 & \textbf{45.6}\\
with metadata uni-\& multi-facet & \textbf{58.6} & \textbf{55.1} &  \textbf{71.8} & \textbf{61.5} & \textbf{64.9}  &\textbf{59.6} & \textbf{56.1} & 40.4\\
\bottomrule
\end{tabular}
}
\caption{Few-Shot Single Turn SQL Prediction Task for Column Ambiguity (Amb) Use Cases Using GPT--4.1(\%). Boldface indicates the comparatively best-performing value}
\end{table*}

\begin{table*}[!h]

\small

\centering
\resizebox{\textwidth}{!}{
\begin{tabular}{l cc cc cc cc}
\toprule
\textbf{Prompting Technique} &
\multicolumn{2}{c}{Unifacet Lexical Amb} &
\multicolumn{2}{c}{Multifacet Lexical Amb} &
\multicolumn{2}{c}{Unifacet Semantic Amb} &
\multicolumn{2}{c}{Multifacet Semantic Amb} \\

\cmidrule(lr){2-3} \cmidrule(lr){4-5} \cmidrule(lr){6-7} \cmidrule(lr){8-9} 
 & LEM & SEM & LEM & SEM  & LEM & SEM & LEM & SEM\\
\midrule
Zero-shot & 33.3 & 22.9 & 48.7 & 5.1 & 34.5 & 24.4& 35.1 & 8.8\\
  without metadata uni-facet & 47.3  &  35.5  &  46.2  &  10.3   & 53.4  &  41.8  &  43.9  &  12.3   \\
 without metadata uni-\& multi-facet  &  49.4  &  38.4   & 41  &  15.4 &   58.3  &  45.9  &  49.1   & 19.3\\
with metadata uni-facet & 59.8  & \textbf{54.7}  &  \textbf{71.8}  &  59  &  61.8  &  57   & 50.9 &   47.4\\
with metadata uni-\& multi-facet & \textbf{60.2}  &  \textbf{54.7 } &  \textbf{71.8}  &   \textbf{61.5 } &  \textbf{64}  &  \textbf{60.3}  &  \textbf{58}  &  \textbf{54.4}\\
\bottomrule
\end{tabular}
}
\caption{Few-Shot Single Turn SQL Prediction Task for Column Ambiguity (Amb) Use Cases Using Grok-3 Mini Fast  (\%). Boldface indicates the comparatively best-performing value}
\end{table*}

\begin{table*}[!h]

\small

\centering
\resizebox{\textwidth}{!}{
\begin{tabular}{l cc cc cc cc}
\toprule
\textbf{Prompting Technique} &
\multicolumn{2}{c}{Unifacet Lexical Amb} &
\multicolumn{2}{c}{Multifacet Lexical Amb} &
\multicolumn{2}{c}{Unifacet Semantic Amb} &
\multicolumn{2}{c}{Multifacet Semantic Amb} \\

\cmidrule(lr){2-3} \cmidrule(lr){4-5} \cmidrule(lr){6-7} \cmidrule(lr){8-9} 
 & LEM & SEM & LEM & SEM  & LEM & SEM & LEM & SEM\\
\midrule
Zero-shot & 31.2 & 21.8 & 18 & 2.6 & 33.8 & 25.2 &26.3 & 8.8\\
  without metadata uni-facet & 52.5  & 36.5  & 61.5  & 10.3  & 58.7  & 44.6  & 52.6   &19.3 \\
 without metadata uni-\& multi-facet  &  53.9  & 40.2  & 59  & 20.5  & 62.4  & 49.5  & 49.1  & 17.5\\
with metadata uni-facet & 60.6 &  56.9  & 59  & 46.1  &  65.8  & 61.3  & 50.9   &\textbf{40.4}\\
with metadata uni-\& multi-facet & \textbf{63.3}   &\textbf{59.4}  & \textbf{69}  & \textbf{48.7}  & \textbf{67.3 } &  \textbf{63.2}  & \textbf{56.1}  & 35.1\\
\bottomrule
\end{tabular}
}
\caption{Few-Shot Single Turn SQL Prediction Task for Column Ambiguity (Amb) Use Cases Using LLaMA-3.3 (\%). Boldface indicates the comparatively best-performing value}
\end{table*}

\begin{table*}[!h]

\small
\centering
\resizebox{\textwidth}{!}{
\begin{tabular}{l cc cc cc cc cc cc cc cc cc cc}
\toprule
\textbf{Model} &
\multicolumn{8}{c}{A/U Use Cases} &
\multicolumn{4}{c}{Conversation Types} \\

\cmidrule(lr){2-9} \cmidrule(lr){10-13}
&{U Lx} &
{M Lx} &
{U Sm} &
{M Sm} &
{U Col Unans} &
{M Col Unans} &
{U Val Amb} &
{M Val Amb} &
{Concise} & {Verbose} & {Partially} & {Not}
\\
\midrule

GPT-4o & 62.8 & 70.8 &55.6   &53.6  & 94    &99   & 39.8 & 43.2 & 67 &  47 &  39.5&  57\\
GPT-4.1 &   64.8 &  72.4 &  62.2 &  65.6   & 87  & 98   &42.1  & 47.6 & 65.2   & 47.3  &  43.4&    63.8\\
GPT-5 &   71.8 & 67.6 & 64  &  65  & 60  & 83    &45.1  & 52  &54.9  & 53.9  & 51.5&   60.4 \\
Grok-3 Mini Fast &    66.2  & 66.2  & 61.8  & 66.8 &  64 &  82  &  44.1  & 53.8 & 54.3  & 48.2  & 44.8 &   65.1\\
LLaMA-3.3 &   66 &  64.4  & 65.6  & 52 &  19 & 47   & 38.9 & 43  & 30.6  & 45.7   &42.7  & 62.5\\
\bottomrule
\end{tabular}
}
\caption{Multi Turn SQL Prediction Task Execution Macro Accuracy (EA-\%) for BIRD Dataset; U - Uni-facet; M - Multi-facet; Col -  Column; Val - Value; Amb - Ambiguity; Unans - Unanswerable; Lx - Lexical Column Amb; Sm - Semantic Column Amb; }
\label{tab:conv_results_bird}
\end{table*}


\begin{table*}[!h]
\centering
\small
\begin{tabular}{l cc cc cc cc cc cc cc}
\toprule
\textbf{Dataset} &
\textbf{Method} &
\textbf{U Col Amb} &
\textbf{M Col Amb} &
\textbf{U Val Amb} &
\textbf{M Val Amb} 
\\

\midrule

\multirow{2}{*}{Spider} & AmbiSQL & (44.8, 69.5) & (12.5, 29.2) & (15, 29.2) & (23.8, 51.1)  \\
 & AmbiSQL-CT  & (59.3, 64.9) & (7.3, 26.8) & (13.1, 27.9) & (34, 48.4) \\

 \multirow{2}{*}{BIRD} & AmbiSQL & (49.7, 59.4) & (0, 9) & (34.6, 40.9) &  (16.3, 40.8)  \\
 & AmbiSQL-CT  & (52.2, 61.9) & (0, 15.5) & (36.5, 46.1) & (20.4, 46.9)   \\

\bottomrule
\end{tabular}
\caption{
Confidence intervals for ambiguity detection task performance (DA\%) in \cref{tab:ad_gpt_col,tab:ad_gpt_val}. Confidence intervals are presented as (lower limit, upper limit) and are computed via bootstrapping with replacement over 10,000 resamples.}

\end{table*}


\begin{table*}[!h]
\centering
\scriptsize
\begin{tabular}{l cc cc cc cc cc cc cc}
\toprule
\textbf{Dataset} &
\textbf{Method} &
\multicolumn{2}{c}{U Lex Amb} &
\multicolumn{2}{c}{M Lex Amb} &
\multicolumn{2}{c}{U Sem Amb} &
\multicolumn{2}{c}{M Sem Amb} &
\multicolumn{2}{c}{U Val Amb} &
\multicolumn{2}{c}{M Val Amb} 
\\

\cmidrule(lr){3-4} \cmidrule(lr){5-6} \cmidrule(lr){7-8} \cmidrule(lr){9-10} \cmidrule(lr){11-12} \cmidrule(lr){13-14} 
& & DA & MA & DA & MA & DA & MA & DA & MA & DA & MA & DA & MA    \\
\midrule

\multirow{2}{*}{Spider} & AmbiSQL & 99.4&  58.6&  \textbf{100}& \textbf{15.4}&  99.7&  56.9&  \textbf{100}&  25.4&  98&  \textbf{22}&  96.8&  35.1  \\
 & \makecell{AmbiSQL - \textsc{Clarity} \\  Exemplars}  & \textbf{99.8} & \textbf{64.3} & \textbf{100} & 12.8 & \textbf{99.8} & \textbf{60} & \textbf{100} & \textbf{26.3} & \textbf{100} & 21.8 & \textbf{100 }& \textbf{40.8}\\

 \multirow{2}{*}{BIRD} & AmbiSQL & 98.6&  55.2&  \textbf{100}&  \textbf{7.7}& \textbf{ 99.2}&  45&  \textbf{100}&  3.1&  \textbf{98.8}&  32.1&   98&  26   \\
 & \makecell{AmbiSQL - \textsc{Clarity} \\  Exemplars}  & \textbf{99.3} & \textbf{60.7} & \textbf{100 }& \textbf{7.7} & \textbf{99.2} & \textbf{54.6} & \textbf{100} & \textbf{6.3} & \textbf{98.8} & \textbf{40.9}& \textbf{ 98} & \textbf{32}  \\

\bottomrule
\end{tabular}
\caption{Ambiguity Detection Task Performance (\%) Using GPT-5. Boldface indicates the comparatively best-performing value}
\label{tab:ad_gpt5_casewise}
\end{table*}

\begin{table*}[!h]
\centering
\scriptsize
\begin{tabular}{l cc cc cc cc cc cc cc}
\toprule
\textbf{Dataset} &
\textbf{Method} &
\multicolumn{2}{c}{U Lex Amb} &
\multicolumn{2}{c}{M Lex Amb} &
\multicolumn{2}{c}{U Sem Amb} &
\multicolumn{2}{c}{M Sem Amb} &
\multicolumn{2}{c}{U Val Amb} &
\multicolumn{2}{c}{M Val Amb} 
\\

\cmidrule(lr){3-4} \cmidrule(lr){5-6} \cmidrule(lr){7-8} \cmidrule(lr){9-10} \cmidrule(lr){11-12} \cmidrule(lr){13-14} 
& & DA & MA & DA & MA & DA & MA & DA & MA & DA & MA & DA & MA    \\
\midrule

\multirow{2}{*}{Spider} & AmbiSQL &  89&49.7  & 93.6 &  33.3 &94& 61.6 & 99.1&  45.6& 98  &14&  97.4&  40.7  \\
 & \makecell{AmbiSQL - \textsc{Clarity} \\  Exemplars}  & 93.1 & 50.8 & 94.9 & 41 & 96.2 & 59 & 100 & 52.6 &  98.4 &  2.5 &  95.4 &  38.3\\

 \multirow{2}{*}{BIRD} & AmbiSQL & 94.5&  40&  100&   7.7&  96.2&  33.9&  96.9&  18.8&    96&  27.3&  96&  24   \\
 & \makecell{AmbiSQL - \textsc{Clarity} \\  Exemplars}  & 94.5 &  43.5 &  100  & 30.8&   98.8 &  44.2 &  100 &  28.1 &  95.7 &  28.3 &  98 &  24 \\

\bottomrule
\end{tabular}
\caption{Ambiguity Detection Task Performance (\%) Using GPT-4o.}
\label{tab:ad_gpt4o_casewise}
\end{table*}

\begin{table}[H]
\centering
\scriptsize
\begin{tabular}{l cc cc cc cc cc cc cc}
\toprule
\textbf{Dataset} &
\textbf{Method} &
\multicolumn{2}{c}{U Lex Amb} &
\multicolumn{2}{c}{M Lex Amb} &
\multicolumn{2}{c}{U Sem Amb} &
\multicolumn{2}{c}{M Sem Amb} &
\multicolumn{2}{c}{U Val Amb} &
\multicolumn{2}{c}{M Val Amb} 
\\

\cmidrule(lr){3-4} \cmidrule(lr){5-6} \cmidrule(lr){7-8} \cmidrule(lr){9-10} \cmidrule(lr){11-12} \cmidrule(lr){13-14} 
& & DA & MA & DA & MA & DA & MA & DA & MA & DA & MA & DA & MA    \\
\midrule

\multirow{2}{*}{Spider} & AmbiSQL &  84.6 & 46.6&  100&  38.5&  94 & 61.8&  97.4&  48.2&  98 & 31 & 84.6 & 33.7 \\
 & \makecell{AmbiSQL - \textsc{Clarity} \\  Exemplars}  & 88.2 &  53.9&   94.9 &  46.2 &  95.8 &  73.2&   100&   63.2&   99.2 &  31.1 &  88.8&   57.7 \\

 \multirow{2}{*}{BIRD} & AmbiSQL & 91&  49&  100&  38.5&  98.9&  46.5&   96.9&  18.8&   93.1&  36.1&  94&  46   \\
 & \makecell{AmbiSQL - \textsc{Clarity} \\  Exemplars}  & 95.9 &  57.9 &   100 &  46.2  & 98.5&   56.2 &  100 &  25 &  96.4 &  43 &  98 &  62 \\

\bottomrule
\end{tabular}
\caption{Ambiguity Detection Task Performance (\%) Using GPT-4.1.}
\label{tab:ad_gpt41_casewise}
\end{table}

\begin{table}[H]
\small
\scriptsize
\centering
\resizebox{\textwidth}{!}{
\begin{tabular}{l cc cc cc cc cc cc cc}
\toprule
\textbf{Dataset} &
\textbf{Method} &
\multicolumn{2}{c}{U Lex Amb} &
\multicolumn{2}{c}{M Lx Amb} &
\multicolumn{2}{c}{U Sm Amb} &
\multicolumn{2}{c}{M Sm Amb} &
\multicolumn{2}{c}{U Val Amb} &
\multicolumn{2}{c}{M Val Amb} 
\\

\cmidrule(lr){3-4} \cmidrule(lr){5-6} \cmidrule(lr){7-8} \cmidrule(lr){9-10} \cmidrule(lr){11-12} \cmidrule(lr){13-14} 
& & DA & MA & DA & MA & DA & MA & DA & MA & DA & MA & DA & MA    \\
\midrule

\multirow{2}{*}{Spider} & AmbiSQL & 91.3 & 59.8&  94.9&  28.2&  96.1&  72.6&  100&  24.6&   96&  21&  91.2&  51.2   \\
 & \makecell{AmbiSQL - \textsc{Clarity} \\  Exemplars}  & 89.6 & 63.3 & 89.7 & 15.4 & 96.2 & 76 & 98.3 & 24.6 & 97.5 & 25.4 & 89.3 & 66.3\\

 \multirow{2}{*}{BIRD} & AmbiSQL & 95.2 &  63.5&  92.3&  15.4&  98.9&  58.5&  100 & 6.3 &  93.6&  40.6 & 96 & 36  \\
 & \makecell{AmbiSQL - \textsc{Clarity} \\  Exemplars}  &  91.7 & 62.8 & 100 & 23.1 & 98.5 & 63.8 & 100 & 15.6 & 91 & 40.4 & 98 & 36 \\

\bottomrule
\end{tabular}
}
\caption{Ambiguity Detection Task Performance (\%) Using Grok-3 Mini Fast.}
\label{tab:ad_grok}
\end{table}

\begin{table}[H]
\centering
\scriptsize
\begin{tabular}{l cc cc cc cc cc cc cc}
\toprule
\textbf{Dataset} &
\textbf{Method} &
\multicolumn{2}{c}{U Lex Amb} &
\multicolumn{2}{c}{M Lex Amb} &
\multicolumn{2}{c}{U Sem Amb} &
\multicolumn{2}{c}{M Sem Amb} &
\multicolumn{2}{c}{U Val Amb} &
\multicolumn{2}{c}{M Val Amb} 
\\

\cmidrule(lr){3-4} \cmidrule(lr){5-6} \cmidrule(lr){7-8} \cmidrule(lr){9-10} \cmidrule(lr){11-12} \cmidrule(lr){13-14} 
& & DA & MA & DA & MA & DA & MA & DA & MA & DA & MA & DA & MA    \\
\midrule

\multirow{2}{*}{Spider} & AmbiSQL &  97&  60.9&  98.7&  50&  97.5&  70&  93&  47.34&   87&  21&  94&  43.9 \\
 & \makecell{AmbiSQL - \textsc{Clarity} \\  Exemplars}  & 94.5 & 58 & 100 & 53.9  &95.9 & 70.1  &96.5&   59.6 & 92.6&  18.9 & 96.4  &49 \\

 \multirow{2}{*}{BIRD} & AmbiSQL & 95.2&  61.4 & 100 & 21.9 & 97.3 & 55.8 & 100 & 21.9   & 92.2&  30.6 & 94&  54   \\
 & \makecell{AmbiSQL - \textsc{Clarity} \\  Exemplars}  & 89.7 & 57.2&  100  &15.4 & 97.3  &54.2 & 96.9&  25 & 95.2 & 40.6 & 96 & 50 \\

\bottomrule
\end{tabular}
\caption{Ambiguity Detection Task Performance (\%) Using LLaMA-3.3.}
\label{tab:ad_llama_casewise}
\end{table}

\end{document}